\documentclass{article}

\usepackage[nonatbib,preprint]{neurips_2022}

\usepackage[utf8]{inputenc} % allow utf-8 input
\usepackage[T1]{fontenc}    % use 8-bit T1 fonts
\usepackage[pagebackref,breaklinks,colorlinks]{hyperref}
\usepackage{url}            % simple URL typesetting
\usepackage{booktabs}       % professional-quality tables
\usepackage{amsfonts}       % blackboard math symbols
\usepackage{nicefrac}       % compact symbols for 1/2, etc.
\usepackage{microtype}      % microtypography
\usepackage[table]{xcolor}         % colors
\usepackage{wrapfig}
\newcommand{\specialcell}[2][c]{%
  \begin{tabular}[#1]{@{}c@{}}#2\end{tabular}}
\newcommand{\specialcelll}[2][l]{%
  \begin{tabular}[#1]{@{}l@{}}#2\end{tabular}}
\definecolor{LightCyan}{rgb}{0.9059,0.9961,1}
\usepackage{multirow}
\usepackage{graphicx}
\usepackage[font=small,labelfont=bf]{caption}  % set global caption size
\usepackage{makecell}
\usepackage{tabulary}
\definecolor{demphcolor}{RGB}{144,144,144}
\newcommand{\demph}[1]{\textcolor{demphcolor}{#1}}
\definecolor{mygray}{gray}{0.4}
\usepackage{pifont}% http://ctan.org/pkg/pifont for xmark and cmark
\newlength\savewidth\newcommand\shline{\noalign{\global\savewidth\arrayrulewidth
  \global\arrayrulewidth 1pt}\hline\noalign{\global\arrayrulewidth\savewidth}}

\newcommand{\tablestyle}[2]{\setlength{\tabcolsep}{#1}\renewcommand{\arraystretch}{#2}\centering\footnotesize}
\makeatletter\renewcommand\paragraph{\@startsection{paragraph}{4}{\z@}
  {.5em \@plus1ex \@minus.2ex}{-.5em}{\normalfont\normalsize\bfseries}}\makeatother
\newcolumntype{C}[1]{>{\centering\arraybackslash}p{#1}}
\newcolumntype{R}[1]{>{\raggedleft\arraybackslash}p{#1}}
\newcolumntype{L}[1]{>{\raggedright\arraybackslash}p{#1}}

\usepackage{etoolbox}
\usepackage{subcaption}
\newdimen\abovecrulesep
\newdimen\belowcrulesep
\makeatletter
\patchcmd{\@@@cmidrule}{\aboverulesep}{\abovecrulesep}{}{}
\patchcmd{\@xcmidrule}{\belowrulesep}{\belowcrulesep}{}{}
\makeatother

\newcommand{\Paragraph}[1]{\vspace{0mm} \noindent \textbf{#1} \hspace{0mm}}
\newcommand{\Section}[1]{\vspace{-1mm} \section{#1} \vspace{-1mm}}
\newcommand{\SubSection}[1]{\vspace{-1mm} \subsection{#1} \vspace{-1mm}}

\definecolor{asparagus}{rgb}{0.53, 0.66, 0.42}
\definecolor{blue(munsell)}{rgb}{0.0, 0.5, 0.69}
\usepackage{xspace}
\newcommand{\modelname}{\textsc{Lavender}\xspace}
\newcommand{\modelbaseline}{\textsc{Lavender-ts}\xspace}
\usepackage{enumitem}
\usepackage{amsmath,amsfonts,bm}

\def\eqref#1{equation~\ref{#1}}
\def\1{\bm{1}}

\def\vv{{\bm{v}}}
\def\vw{{\bm{w}}}

\DeclareMathAlphabet{\mathsfit}{\encodingdefault}{\sfdefault}{m}{sl}
\SetMathAlphabet{\mathsfit}{bold}{\encodingdefault}{\sfdefault}{bx}{n}

\title{\modelname
: Unifying Video-Language Understanding as Masked Language Modeling}

\author{{\bf Linjie Li, Zhe Gan, Kevin Lin, Chung-Ching Lin,  Zicheng Liu, Ce Liu, Lijuan Wang}\\
Microsoft\\
{\tt\small \{linjli,zhgan,keli,chungching.lin,zliu,ce.liu,lijuanw\}@microsoft.com}}

\begin{document}

\maketitle
\begin{abstract}
Unified vision-language frameworks have greatly advanced in recent years, most of which adopt an encoder-decoder architecture to unify image-text tasks as sequence-to-sequence generation. However, existing video-language (VidL) models still require task-specific designs in model architecture and training objectives for each task. In this work, we explore a unified VidL framework \modelname, where Masked Language Modeling~\cite{devlin2019bert} (MLM) is used as the common interface for all pre-training and downstream tasks. 
% in which all pre-training and downstream tasks are formulated as Masked Language Modeling~\cite{devlin2019bert} (MLM), without any task-specific instructions or layers. 
Such unification leads to a simplified model architecture, where only a \textit{lightweight} MLM head, instead of a  decoder with much more parameters, is needed on top of the multimodal encoder. Surprisingly, experimental results show that this unified framework achieves competitive performance on 14 VidL benchmarks, covering video question answering, text-to-video retrieval and video captioning. Extensive analyses further demonstrate the advantage of \modelname over existing VidL methods in: ($i$) supporting all downstream tasks with just a single set of parameter values when multi-task finetuned; ($ii$) few-shot generalization on various downstream tasks; and ($iii$) enabling zero-shot evaluation on video question answering tasks. Code is available at \url{https://github.com/microsoft/LAVENDER}.
\end{abstract}

\begin{figure}[h]
    \centering
    \vspace{-2ex}
    \includegraphics[width=.97\textwidth]{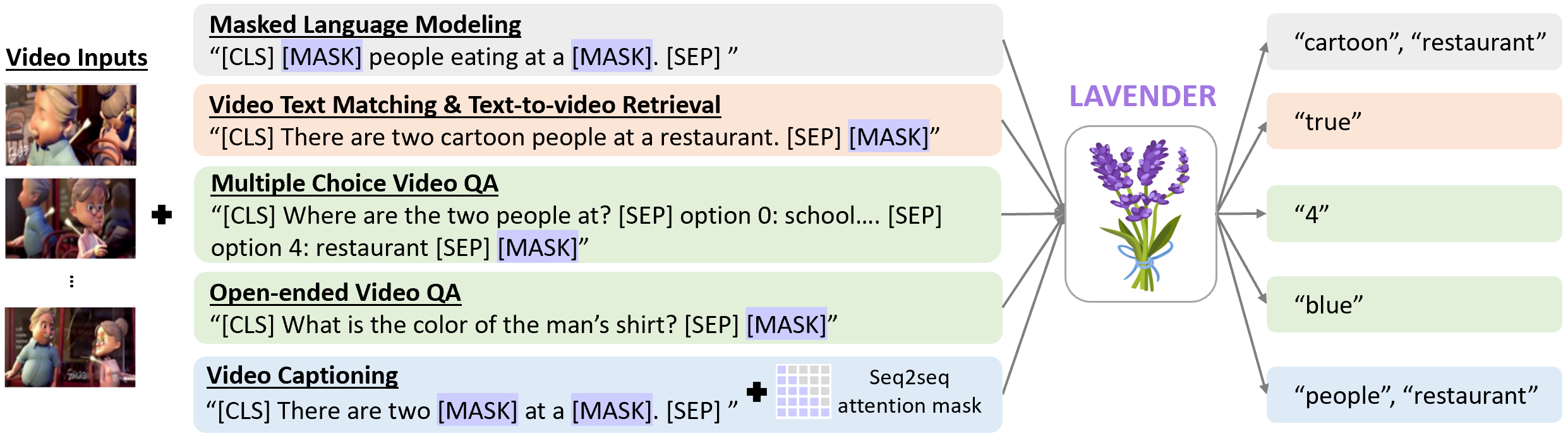}
    \vspace{-1ex}
    \caption{Overview of \modelname (\textbf{LA}nguage-\textbf{V}id\textbf{E}o u\textbf{NDER}standing) model. \modelname unifies both pre-training and downstream finetuning as Masked Language Modeling.
    }
    \vspace{-2ex}
    \label{fig:lavender_framework}
\end{figure}

\Section{Introduction}\label{sec:intro}
Large-scale transformer-based pre-training is now the \emph{de facto} practice for NLP and vision-language research~\cite{devlin2019bert,liu2019roberta,raffel2020exploring,radford2021clip,jia2021scale-up}. Together with the great success of image-text pre-training~\cite{tan2019lxmert,lu2019vilbert,chen2020uniter,li2020oscar}, video-language (VidL) pre-training~\cite{sun2019videobert,zhu2020act-bert,li2020hero,lei2021clip-bert,zellers2021merlot} has also received an increasing amount of attention. By pre-training an end-to-end multimodal transformer on a large number of video-text pairs, state-of-the-art performance has been achieved across a wide range of VidL tasks, including video question answering (QA)~\cite{jang2017tgif-qa,xu2017msrvtt-msvd-qa}, text-to-video retrieval~\cite{hendricks2017didemo,rohrbach2015lsmdc}, and video captioning~\cite{xu2016msrvtt,wang2019vatex}. These advances are encouraging; however, on the other hand, all existing VidL works require designing task-specific heads on top of the transformer encoder for each pre-training or downstream task. For example, during pre-training, separate Masked Language Modeling ~\cite{devlin2019bert} (MLM) and Video Text Matching (VTM) heads are used, while a new, separately parameterized head needs to be added for each downstream adaptation. Furthermore, due to the particular nature of different tasks, they are typically modeled using different training objectives. For example, multiple-choice video QA is formulated as a classification problem, while video captioning is inherently a generation task. A natural but challenging question arises: \emph{can we have a unified architecture that supports all the popular VidL tasks simultaneously without introducing task-specific heads?}

To answer this, we present \modelname, a unified VidL framework where all pre-training and downstream tasks are formulated as simple MLM. As shown in Figure~\ref{fig:lavender_framework}, we use two pre-training tasks: MLM and VTM. However, for VTM, instead of adding a head on top of the output of the commonly used \texttt{[CLS]} token, as used in all existing works, we propose to append the same \texttt{[MASK]} token that is used for MLM at the end of the video-text input, and use the same MLM head to predict whether the input video-text pair matches or not. Note that VTM is typically formulated as a binary classification problem; here, we simply treat the output of \texttt{true} or \texttt{false} from VTM as natural language tokens directly predicted from the whole vocabulary, so that the same set of parameters can be used for both MLM and VTM. 

During downstream adaptation, instead of discarding the MLM head used in pre-training and adding new heads, which is the standard practice for all previous VidL works, we use the same MLM head used in pre-training for all downstream tasks. Specifically,
\vspace{-1mm}
\begin{itemize}[leftmargin=*]
\setlength\itemsep{0.1mm}
    \item For text-to-video retrieval, we train the model in the same way as in the VTM pre-training task. During inference, for each text query, we concatenate it with each candidate video, and calculate the corresponding probability of the \texttt{[MASK]} token being predicted as \texttt{true}, and then rank all candidate videos based on that score. 
    \item For multiple-choice video QA, we concatenate the question and each answer candidate sequentially, and add a \texttt{[MASK]} token at the end of the sequence, and use the same MLM head to predict the answer as ``\texttt{n}'' (assuming the ground-truth choice is the \texttt{n}-th answer).    
    \item For open-ended video QA, since most of the ground-truth answers in our tested datasets only contain one word, we simply append a \texttt{[MASK]} token to the end of the video-question input, and let the model predict the answer from the whole vocabulary.         
    \item For video captioning, during training, we mask a certain percentage of the tokens, and then predict the masked tokens using a seq2seq attention mask~\cite{li2020oscar,zhang2021vinvl}. During inference, the full caption is auto-regressively predicted, by inserting \texttt{[MASK]} tokens one at a time. 
\end{itemize}

\begin{table}[t]
\centering
    \tablestyle{3pt}{1.1}
    \caption{New state-of-the-art performance with \modelname~across 12 video-language understanding tasks. Accuracy, average(R1, R5, R10) and CIDEr scores are reported for video QA, retrieval and captioning.}
    \vspace{1ex}
    \label{tab:sota}
    \scriptsize
    \resizebox{\linewidth}{!}{
    \begin{tabular}{ccccccccccccc} 
    \shline
    & \multicolumn{3}{c}{\scriptsize{TGIF}} & \multicolumn{2}{c}{\scriptsize{MSRVTT}} &  \multicolumn{2}{c}{\scriptsize{LSMDC}} & \scriptsize{MSVD} &  \multicolumn{2}{c}{\scriptsize{Captioning}} & \multicolumn{2}{c}{\scriptsize{Retrieval}}\\
    \cmidrule(lr){2-4} \cmidrule(lr){5-6} \cmidrule(lr){7-8} \cmidrule(lr){9-9}  \cmidrule(lr){10-11} \cmidrule(lr){12-13}
    & \scriptsize{Action}
    & \scriptsize{Transition}
    & \scriptsize{Frame}
    & \scriptsize{MC}
    & \scriptsize{QA}
    & \scriptsize{MC}
    & \scriptsize{FiB}
    & \scriptsize{QA} 
    & \scriptsize{MSRVTT}
    & \scriptsize{MSVD}& \scriptsize{DiDeMo}
    & \scriptsize{LSMDC}\\
    \shline
    \multirow{2}{*}{\specialcell{\scriptsize{Published}\\\scriptsize{SOTA}}} &  94.0  & 96.2   & 69.5   &  90.9 & 43.1  & 81.7  & 52.9   & 46.3 &  60.0 & 120.6  &    65.1     & 41.9 \\
    & \cite{zellers2021merlot} % TGIF-Action
    & \cite{zellers2021merlot} % TGIF-Transition
    & \cite{zellers2021merlot}   % TGIF-Frame
    & \cite{zellers2021merlot} % MSRVTT-MC
    
    & \cite{zellers2021merlot} % MSRVTT-QA
    & \cite{zellers2021merlot} % LSMDC-MC
    & \cite{zellers2021merlot} % LSMDC-FiB
    & \cite{yang2021just-ask} % MSVD-QA
    & \cite{MV-GPT} % msrvtt
    & \cite{lin2021swinbert} % msvd
    & \cite{qb-norm} % didemo
    & \cite{camoe} % lsmdc
    \\
    \hline
    \scriptsize{\modelname}          & 96.6   & 99.1 & 73.5   & 97.4 & 45.0   & 87.0   & 57.1  &   56.6      & 60.1 &  150.7   & 72.4   &       43.3 \\
    % \midrule
    $\Delta$                    & 2.6$\uparrow$     & 2.9$\uparrow$     & 4.0$\uparrow$  & 6.5$\uparrow$    & 1.9$\uparrow$    &  5.3$\uparrow$     & 4.2$\uparrow$    & 10.3$\uparrow$            & 0.1$\uparrow$   & 30.1$\uparrow$     & 7.3$\uparrow$   &      1.4$\uparrow$    \\
    \shline
    \end{tabular}
    }
    \vspace{-4ex}
\end{table}
\modelname is inspired by VL-T5~\cite{VL-T5}, UNICORN~\cite{unicorn} and OFA~\cite{OFA} that aim to provide a unified pre-training framework for image-text tasks. However, ours is very different from theirs, as we use an encoder-only model and an additional lightweight MLM head on top of it, while a heavy transformer decoder is needed in~\cite{VL-T5,unicorn,OFA}. By unifying all the VidL tasks as MLM, \modelname can seamlessly adapt to different VidL tasks, and enables new capabilities over existing task-specific methods, such as ($i$) supporting different VidL tasks with a single set of parameter values when multi-task finetuned; ($ii$) better generalizability to test data under few-shot finetuning; and ($iii$) zero-shot inference on video question answering. Surprisingly, by using this simple generative approach, we outperform previously published state-of-the-arts on 12 out of 14 downstream tasks (Table~\ref{tab:sota}), even when pre-trained with much fewer data (see Section~\ref{sec:sota} for detailed comparisons). 
% \vspace{-1ex}
\Section{Related Work}
\Paragraph{Video-Language Pre-training.} 
Branching out from large-scale image-text pre-training~\cite{chen2020uniter,tan2019lxmert}, researchers have been leveraging large-scale multimodal data~\cite{chen2015coco,vg,sharma2018cc,miech2019howto100m,zellers2021merlot,bain2021frozen,zellers2022merlot} to build pre-trained video-language (VidL) models~\cite{luo2020univl,patrick2021support-set,akbari2021vatt,yang2021taco,xu2021video-clip,rouditchenko2021avlnet,xu2021vlm} for a wide range of generative~\cite{zhou2018youcook2,xu2016msrvtt,lei2020tvr} and discriminative~\cite{jang2017tgif-qa,lsmdc-fib,lei2018tvqa} tasks. Prominent examples include VideoBERT~\cite{sun2019videobert}, HERO~\cite{li2020hero}, ActBERT~\cite{zhu2020act-bert}, ClipBERT~\cite{lei2021clip-bert} and MERLOT~\cite{zellers2021merlot}.
Popular pre-training tasks include Masked Language Modeling (MLM)~\cite{su2020vlbert}, Video Text Matching (VTM)~\cite{lei2021clip-bert}, frame order modeling~\cite{li2020hero,zellers2021merlot} and masked visual modeling~\cite{li2020hero,fu2021violet}. For simplicity, we only use MLM and VTM losses for pre-training, as in our experiments, we observe that the other pre-training tasks are not that essential to improving the final performance on downstream tasks (results in Appendix~\ref{app:res}).

Although achieving strong performance, existing methods all require task-specific architectures or objectives for different downstream tasks. For example, text-to-video retrieval~\cite{hendricks2017didemo,rohrbach2015lsmdc} is modeled as binary classification~\cite{lei2021clip-bert} or via contrastive learning~\cite{miech2020end,gabeur2020mmt}; video question answering~\cite{jang2017tgif-qa,xu2017msrvtt-msvd-qa} is often formulated as multi-class classification with a set of pre-defined answer candidates~\cite{zellers2021merlot,siamese}; and video captioning can be tackled via MLM with a multi-layer perceptron~\cite{lin2021swinbert} or prefix language modeling with a text decoder~\cite{MV-GPT}.

\Paragraph{Unified Frameworks for Multimodal Understanding.}
There have been attempts in building an omnipotent model that can simultaneously handle different tasks with a unified architecture, which can be broadly categorized into two directions. The first is to insert expert-designed task-specific heads for each downstream task~\cite{GPV,FLAVA,UniT,li2021value}. 
These task-specific output layers not only require expert knowledge, but are also unlikely to generalize to new tasks. For example, when a new question answering task comes in, a new fully-connected layer with output dimension of answer vocabulary size is required. The second direction is to unify the input-output format of different downstream tasks ~\cite{simvlm,VL-T5,unicorn,OFA}. With a unified vocabulary, different downstream tasks (\textit{e.g.,} image question answering~\cite{balanced_vqa_v2}, image captioning~\cite{chen2015coco} and visual grounding~\cite{referit}) can be formulated as the sequence-to-sequence generation with the shared encoder-decoder architecture~\cite{unicorn,OFA}. 

\begin{figure*}[t!]
    \centering
    \begin{subfigure}[t]{0.28\textwidth}
        \centering
        \includegraphics[width=\textwidth]{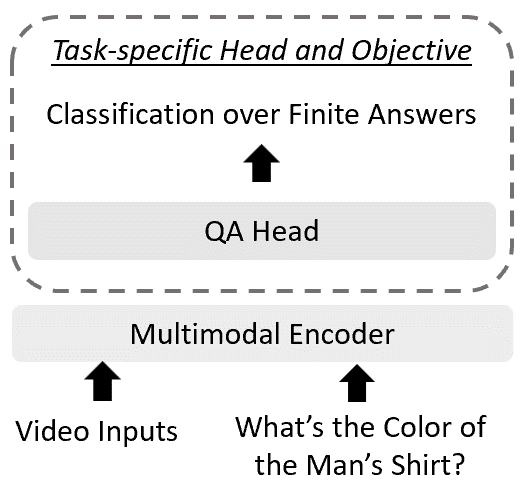}
        \caption{Existing VidL Methods}
        \label{subfig:existing}
    \end{subfigure}%
    \hfill 
    \begin{subfigure}[t]{0.38\textwidth}
        \centering
        \includegraphics[width=0.73\textwidth]{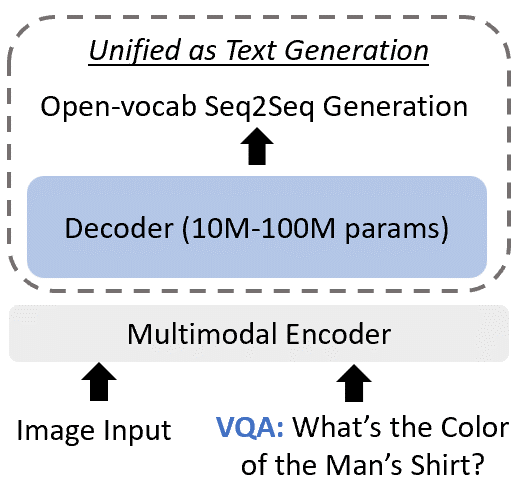}
        \caption{\footnotesize Existing Unified Image-text Models}
        \label{subfig:vl-t5}
    \end{subfigure}%
    \hfill 
    \begin{subfigure}[t]{0.28\textwidth}
        \centering
        \includegraphics[width=\textwidth]{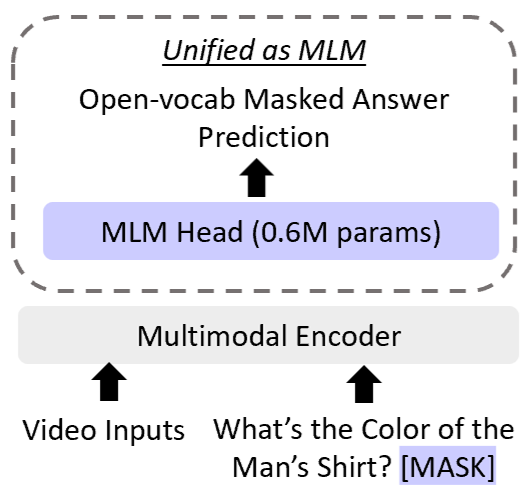}
        \caption{\modelname}
        \label{subfig:ours}
    \end{subfigure}%
    \vspace{-1ex}
    \caption{Illustration of the \textbf{differences between \modelname and existing methods} with image/video question answering (QA) as an example. Unlike task-specific designs in existing VidL methods, \modelname unifies all tasks as MLM (Figure~\ref{fig:lavender_framework}). We adopt an encoder-only architecture, with a lightweight MLM head, instead of the heavy  decoder in unified image-text models (\emph{e.g.}, VL-T5~\cite{VL-T5} with task-specific prefix in text input).
    }
    \label{fig:comparison}
    \vspace{-3ex}
\end{figure*}

\textbf{Our work} aims to provide a unified framework for VidL understanding, in contrast to task-specific architectures and objectives used in existing VidL models (Figure~\ref{subfig:ours} vs. Figure~\ref{subfig:existing}). \textsc{Lavender} differs from the previous unified image-text models in that all pre-training and downstream tasks are unified as MLM, and a simple encoder-only architecture with a lightweight MLM head is used, instead of sequence-to-sequence modeling as in~\cite{VL-T5,simvlm,OFA,unicorn} that also requires a heavy transformer decoder (Figure~\ref{subfig:ours} vs. Figure~\ref{subfig:vl-t5}). 
\Section{\modelname}
\SubSection{Model Architecture}
% A diagram of \modelname architecture is presented in Figure~\ref{subfig:ours}.
Given a pair of text sentence $\{\vw_{n}\}|_{n=1}^{N}$ and a video $\{\vv_{t}\}|_{t=1}^{T}$, we first encode them separately via unimodal encoders (\textit{i.e.}, vision encoder and text encoder) to generate unimodal features. 
Here, $N$ is the number of tokens in a sentence and $T$ is the number of frames sampled from the input video. 
We follow previous works~\cite{lei2021clip-bert,zellers2021merlot} to only sparsely sample a few frames to ease the computational burden. A multimodal fusion encoder (dubbed as fusion encoder) 
% operates on top of these unimodal encoders and 
projects textual features and visual features into a shared embedding space to learn cross-modal representations. As \modelname unifies both pre-training and downstream tasks as Masked Language Modeling (MLM), the same MLM head is used to generate the final outputs from the cross-modal representations, across different tasks. Next, we explain each component in detail.

\Paragraph{Vision Encoder.} Inspired by the success of vision transformers in modeling spatial details in images~\cite{dosovitskiy2021vit,liu2021swin}, different transformer architectures~\cite{timesformer,liu2021video-swin,yan2022multiview} have been proposed to model the long-range temporal modeling in videos, achieving promising results on action recognition~\cite{kay2017kinetics}. Recent video-language works~\cite{fu2021violet,lin2021swinbert} have started to embrace the success of video transformers, demonstrating stronger performance than encoding each video frame independently~\cite{lei2021clip-bert}. 
In our work, we adopt Video Swin Transformer~\cite{liu2021video-swin} (VidSwin) as the vision encoder to encode the raw video frame inputs as a sequence of visual features. Given $T$ input video frames $\{\vv_{t}\}|_{t=1}^{T}$ of size $H\times W\times 3$, we first split each frame into non-overlapping patches of size $h\times w$. VidSwin additionally enforces temporal downsampling of size 2 as a preprocessing step. To allow \modelname to have the flexibility of utilizing both video-text and image-text data for pre-training, we remove this temporal downsampling. As a result, we can extract a sequence of visual features of size $T\times\frac{H}{h}\times\frac{W}{w}$ from the last encoder block of VidSwin. Each feature is of size $8C$ ($C$ is the channel dimension), which is projected to the same dimensional space as text features via a fully-connected layer. We follow~\cite{fu2021violet} to add learnable positional embedding layers along both spatial and temporal dimensions. The resulting visual features are used as input to the fusion encoder to learn cross-modal representations.

\Paragraph{Text Encoder.} The input text sentence is first tokenized into the sequence of word tokens  $\{\vw_{n}\}|_{n=1}^{N}$, following~\cite{wu2016word-piece}. Two special tokens \texttt{[CLS]} and \texttt{[SEP]} are inserted at the beginning and the end of the token sequence. We follow previous works~\cite{lei2021clip-bert,zellers2021merlot,fu2021violet} to adopt a lightweight word embedding layer~\cite{devlin2019bert} as the text encoder. The high-dimensional text embeddings are concatenated with visual features and then fed into the fusion encoder. 

\Paragraph{Multimodal Fusion Encoder.} The fusion encoder is a 12-layer, 768-dimensional Transformer~\cite{vaswani2017attention}, mirroring the BERT-base architecture~\cite{devlin2019bert}. To compute the cross-modal representations, the unimodal features from vision and text encoders are fused together via self-attention operations. 

\SubSection{Our Unified Framework}\label{sec:unify}
Now, we introduce how \modelname can be trained in a unified way.

\Paragraph{Video-language Pre-training.} We adopt two objectives to pre-train \modelname. The first is \textbf{Masked Language Modeling} (MLM), which is directly adopted from language model pre-training~\cite{devlin2019bert,liu2019roberta}. In MLM, we randomly replace 15\% of word tokens with a \texttt{[MASK]} token, a random word, or the same word.  The goal is to reconstruct the correct tokens based on the corresponding hidden representations from the output of the fusion encoder at the masked position. A multi-layer perceptron (MLP) with output dimension as \texttt{vocab\textunderscore size} 
% (\textit{e.g.}, 30,522 in bert-base-uncased) 
projects these hidden representations into the discrete word token space. Cross-entropy loss is used to supervise the model training. The second is \textbf{Video Text Matching}, but reformatted as MLM (VTM as MLM). Specifically, we append a \texttt{[MASK]} token to the textual sentence to mimic the masked textual inputs in MLM. At each training step, we randomly replace the corresponding text for a given video with a text description from a different video in the same batch. At the masked position, \modelname reuses the exact same MLP used in MLM to make a prediction. Although the ground-truth label is restricted to two tokens (\textit{i.e.}, \texttt{true} (\texttt{false}) for a positive (negative) video-text pair), but the model predictions are made across all vocabularies.

\Paragraph{Downstream Adaptation.} As shown in Figure~\ref{fig:lavender_framework}, we can readily apply the pre-trained \modelname to 4 types of downstream tasks, including text-to-video retrieval, multiple-choice video question answering, open-ended video question answering and video captioning. For each task, we transform the text input by inserting or replacing existing tokens with \texttt{[MASK]} tokens, so that all tasks can be supervised with cross-entropy loss, and the final predictions are made based on the word token predicted at the masked position. 
Here, we explain in detail how to construct the masked textual inputs and generate model predictions for each downstream task. 

For \textbf{text-to-video retrieval}, similar to VTM during pre-training, we insert a \texttt{[MASK]} token at the end of the text input. During training, we treat corresponding video-text pairs as positives (with ground-truth label \texttt{true}) and all other pairwise combinations constructed by replacing the ground-truth text with a randomly sampled one as negatives (with ground-truth label \texttt{false}). During inference, given a textual query, we rank the videos according to the model confidence of predicting \texttt{true} at the masked position. 
For \textbf{multiple-choice video QA}, we concatenate each answer choice (\texttt{A$_{\text{n}}$}) sequentially to the question (\texttt{Q}) with a \texttt{[SEP]} token in between. A \texttt{[MASK]} token is then added at the end, to allow the model to make a prediction of the correct index for the ground-truth answer choice. For example, for a question and 5 answer choices, we take \texttt{Q+[SEP]+A$_0$+[SEP]+...+A$_{4}$+[MASK]} as the text input. If \texttt{A$_{\text{n}}$} is the correct answer, the ground-truth label for the masked token is \texttt{n}. Through the MLM head, the model makes a prediction at the masked position over the whole vocabulary. During inference, to ensure a valid answer, we take the most probable predictions over all answer indices (\textit{e.g.}, \{\texttt{0,1,2,3,4}\}).
For \textbf{open-ended video QA}, we similarly inject  \texttt{[MASK]} tokens after the question. For simplicity, we only add one \texttt{[MASK]} token.\footnote{Note that we can optionally insert multiple \texttt{[MASK]} tokens to allow answer predictions with variant lengths. However, as we will show in Appendix~\ref{app:downstream_dataset}, over 95\% of the questions in the evaluation datasets considered can be answered with a single word.} We then tokenize the ground-truth answers as the ground-truth label for masked prediction. If the tokenized answer is longer than 1 word, we simply ignore it during training, and regard it as a wrong prediction during inference.
For \textbf{video captioning}, we use a causal self-attention mask where the caption token can only attend to the existing output tokens, which simulates a uni-directional seq2seq generation process, following~\cite{lin2021swinbert}. During training, we randomly ``mask'' some words with \texttt{[MASK]} token and apply the MLM objective. During inference, the caption is generated in an auto-regressive manner. At each generation step, the model sees the entire video input and previously generated tokens, plus a \texttt{[MASK]} token, at which the model makes a prediction for the current token.

\Section{Experiments}\label{sec:exp}
In this section, we first describe our experimental settings (\ref{sec:exp_settings}), and show the superiority of \modelname over comparable task-specific baselines under both single-task (\ref{sec:task-specific}) and multi-task (\ref{sec:all-in-one}) finetuning settings. We then show that our model can better generalize to testing data under few-shot finetuning and has strong zero-shot capability on video question answering benchmarks (\ref{sec:zs-few-shot}). Lastly, we compare \modelname with prior arts and show we outperform published methods on 12 out of 14 benchmarks, even when pre-trained with much fewer data (\ref{sec:sota}). 

\SubSection{Experimental Settings}\label{sec:exp_settings}

\Paragraph{Pre-training Data.} In our default setting, we follow \cite{bain2021frozen} to aggregate video-text pairs in WebVid2.5M~\cite{bain2021frozen} and image-text pairs in CC3M~\cite{sharma2018cc} to pre-train \modelname.  As a scale-up recipe, we additionally crawl 11.9M video-text pairs from the web, following the same procedure in~\cite{bain2021frozen}. We similarly scale up image-text pairs by assembling COCO~\cite{chen2015coco}, Visual Genome~\cite{vg}, SBU Captions~\cite{sbu}, CC12M~\cite{changpinyo2021cc12m} and CC3M. Combining these video-text and image-text datasets together results in 14M videos + 16M images. Unless otherwise specified, all results reported in this section as \modelname are pre-trained under the default setting with 2.5M videos + 3M images. In Section~\ref{sec:sota}, we show that scaling up our pre-training data further improves model performance. 

\Paragraph{Downstream Tasks.}
We evaluate \modelname on 14 video-language benchmarks over popular VidL tasks, including text-to-video retrieval, video question answering (QA) in both multiple-choice (MC) and open-ended (OE) settings and video captioning. We briefly list the evaluation datasets for each task type below; detailed descriptions are included in Appendix~\ref{app:downstream_dataset}.
\vspace{-2mm}
\begin{itemize}[leftmargin=*]
\setlength\itemsep{0mm}
\item Text-to-video Retrieval: MSRVTT~\cite{xu2016msrvtt}, DiDeMo~\cite{hendricks2017didemo}, MSVD~\cite{chen2011msvd} and LSMDC~\cite{rohrbach2015lsmdc};
\item MC Video QA: TGIF-Action, TGIF-Transition~\cite{jang2017tgif-qa}, MSRVTT-MC~\cite{msrvtt-mc} and LSMDC-MC~\cite{torabi2016lsmdc-mc};
\item OE Video QA: TGIF-Frame~\cite{jang2017tgif-qa}, MSRVTT-QA, MSVD-QA~\cite{xu2017msrvtt-msvd-qa} and LSMDC-FiB~\cite{lsmdc-fib};
\item Video Captioning: MSRVTT~\cite{xu2016msrvtt} and MSVD~\cite{chen2011msvd}.
\end{itemize}
\vspace{-2mm}
\Paragraph{Implementation Details.}
We initialize our Vision Encoder with VideoSwin-Base~\cite{liu2021video-swin}, pre-trained on Kinetics-600~\cite{kay2017kinetics}. Text Encoder and Multimodal Fusion Encoder are initialized from pre-trained BERT-Base~\cite{devlin2019bert}. We train \modelname in an end-to-end manner for both pre-training and downstream finetuning. Our implementation of \modelname is based on PyTorch~\cite{paszke2019pytorch}. We adopt AdamW~\cite{loshchilov2019adamw} as the optimizer with an initial learning rate of 2e-5, betas of (0.9, 0.98), and weight decay of 1e-3 for all pre-training and finetuning experiments. During pre-training, we sparsely sample $T$=4 video frames and resize them into $H$=$W$=224 to split into patches with size $h$=$w$=32. For default setting with 2.5M videos + 3M images, we pre-train \modelname for 10 epochs on 32 NVIDIA V100 GPUs with a batch size of 28 per GPU, which takes about 2 days. The scale-up pre-training takes about 10 days on 64 NVIDIA V100 GPUs. For all downstream tasks, we adopt the same video frame size and patch size, but $T$=5 video frames. 
More details can be found in Appendix~\ref{app:implementation}.
\begin{table}[t]
\centering
    \tablestyle{5pt}{1.1} 
    \def \w{20pt} 
    \caption{\textbf{Comparison to task-specific baseline} under single-task (ST)  and multi-task (MT) finetuning (FT), with or without video-langauge pre-training (PT). We report accuracy for TGIF-Action and MSVD-QA, average (R1, R5, R10) for DiDeMo Retrieval (Ret.) and CIDEr score  for MSRVTT Captioning (Cap.). Meta-Ave. is the average score across all evaluation datasets. P and H denote the total parameter count in the backbone of \modelname (vision encoder + text encoder + fusion encoder) and top output layer.  Note that the baseline with task-specific heads (L3-6) is \modelbaseline, introduced in Section~\ref{sec:task-specific}.
    }
    \label{table:task-specific}
    \vspace{1ex}
    \footnotesize
    % \resizebox{\linewidth}{!}{
    \begin{tabular}{llcc|>{\centering\small}p{0.02\textwidth}|ccccc}
        % \toprule
        
        \shline
          ~ & \footnotesize{Task-specific} & ~ & ~ & ~ & \footnotesize{Meta} & \footnotesize{TGIF}  & \footnotesize{MSVD} & \footnotesize{DiDeMo}  & \footnotesize{MSRVTT}\\
          \cmidrule(lr){7-7} \cmidrule(lr){8-8} \cmidrule(lr){9-9} \cmidrule(lr){10-10}
        \footnotesize{PT} & \footnotesize{designs} & \footnotesize{FT} & \footnotesize{\#Params} & \# & \footnotesize{Ave.} & \footnotesize{Action}  & \footnotesize{QA}  & \footnotesize{Ret.}  & \footnotesize{Cap.} \\
        \shline
        \multirow{4}{*}{-} & \multirow{2}{*}{-} & ST & 4(P+H) & 1 & 45.5  & 93.5 & 40.8 & 
        % Not Converge
        0.0
        & 47.7 \\
        & & MT & P+H & 2 & 58.5 & 95.9 & 47.4 & 
        % 17.0/46.4/60.1 
        41.2
        & 50.0\\
        \cline{2-10}
        & \multirow{2}{*}{Head} & ST & 4(P+H) & 3 &  40.1 & 31.9 & 44.2 & 
        % 15.8/40.4/53.8
        36.7
        & 47.4\\
        & & MT & P+4H & 4 & 55.6 & 94.1 & 44.6 & 
        % 14.2/38.4/53.5
        35.4
        & 48.3\\
        \hline
        \multirow{2}{*}{\specialcelll{VTM+MLM}} & \multirow{2}{*}{Head} & ST & 4(P+H) & 5 & 64.0 & 94.5 & 46.7 & 
        % 36.2/64.5/76.2 
        59.0
        & 55.7\\
        & & MT & P+4H & 6 & 62.4 & 95.5 & 47.7 & 
        % 29.6/58.2/71.3 
        53.0
        & 53.3\\
        \hline
        \multirow{4}{*}{\specialcelll{VTM (as MLM)\\+MLM}} & - & ST & 4(P+H) & 7 & \textbf{68.9} & 95.8 & \textbf{54.4} & 
        % \textbf{46.4}/\textbf{74.4}/82.6 
        \textbf{68.2}
        & 57.3\\
        \cline{2-10}
        & - & \multirow{3}{*}{MT} & \multirow{3}{*}{P+H} & 8 & 68.3 & \textbf{96.5} & 53.5 & 
        % 43.1/71.2/\textbf{83.2} 
        65.8
        & \textbf{57.4} \\
        & Task Prompt &  & & 9 & 67.9 & 96.2 & 53.4 & 
        % 43.1/71.2/82.6 
        65.6
        & 56.4 \\
        & Task Token &  & & 10 & 67.9 & \textbf{96.5} & 53.6 & 
        % 43.5/70.1/81.2 
        64.9
        & 56.7 \\
        \shline
    \end{tabular}
    \vspace{-3ex}
\end{table}

\SubSection{Comparison to Task-specific Baseline}\label{sec:task-specific}
To make a fair comparison to task-specific methods, we train a task-specific version of \modelname (denoted as \modelbaseline). We replace the shared Masked Language Modeling (MLM) head in \modelname with task-specific heads and adopt task-specific objectives. 
% to train the model. 
For \textbf{text-to-video retrieval} (and similarly for video text matching during pre-training), a multi-layer perceptron (MLP) with output dimension 1 is applied over the global video-language representation of the \texttt{[CLS]} token and binary cross-entropy loss is adopted to supervise the model training. For \textbf{multiple-choice video question answering} (QA), we concatenate questions with all answer candidates to form the input text, similar to what was described in Section~\ref{sec:unify}, but without the added \texttt{[MASK]} token. A task-specific MLP with output dimension as the number of answer choices is applied over the representation of \texttt{[CLS]} token and cross-entropy loss is used to train a classifier over all answer indices (\textit{e.g.,} $\{0,1,2,3,4\}$ with 5 answer choices). For \textbf{open-ended video QA}, we follow the common practice~\cite{zellers2021merlot,fu2021violet} to build a finite set of answer vocabularies covering the most common answers in the training split of each dataset. Similarly, a MLP with output dimension as the number of answers is added and cross-entropy loss is used to train a classifier over all answers. For \textbf{video captioning}, we simply adopt the same training strategy with MLM head as in our unified model.

Table~\ref{table:task-specific} compares \modelname to the task-specific baseline \modelbaseline under different settings, on four representative benchmarks. For easier comparisons, we use Meta-Ave, the average across scores over all evaluation tasks, to measure the average model performance. Here, we focus our discussion on results under single-task finetuning, and delay the analysis on multi-task finetuning to Section~\ref{sec:all-in-one}.
Without video-language (VidL) pre-training, task-specific baseline with different task heads (L3) outperforms our unified model (L1) on MSVD-QA and DiDeMo Retrieval. Captioning performance are similar as we adopt the same MLM head and finetuning strategy for both models. The only outlier is TGIF-Action, on which we empirically find the training of \modelbaseline struggles to converge, leading to a low Meta-Ave score.

We also compare the two models under VidL pre-training. We follow~\cite{lei2021clip-bert} to pre-train task-specific \modelbaseline with MLM and the standard Video Text Matching (VTM) task, which is modeled as binary classification with an additional MLP layer. In comparison, the unified model \modelname is pre-trained with MLM and VTM as MLM, with the shared MLM head (Section~\ref{sec:unify}). The unified VidL pre-training (L7) significantly enhances the performance of \modelname, with a gain of +23.4 on Meta-Ave over without pre-training (L1). 
Comparing both models under VidL pre-training, we also observe that \modelname (L7) outperforms \modelbaseline (L5) by a notable margin across all 4 tasks (+4.9 on Meta-Ave). 

\SubSection{Multi-task Finetuning}\label{sec:all-in-one}
In this section, we aim to answer the important question raised in Section~\ref{sec:intro}:  \textit{can we have a unified architecture that supports all downstream tasks simultaneously without introducing task-specific heads?} We first compare \modelname with several multi-task learning baselines with different task-specific designs in Table~\ref{table:task-specific} and report results under the \textit{extreme} multi-task setting - a single model that can support all 14 datasets (all-in-one) in Table~\ref{table:multi-task}.

\Paragraph{Comparison to task-specific baseline.} We begin our comparison with the most common multi-task baseline in literature~\cite{li2021value,lu202012in1} - adopting a task-specific head and objective for each task, while sharing the backbone. This is equivalent to finetuning the task-specific model \modelbaseline under multi-task setting.
We compare the two models in Table~\ref{table:task-specific} and summarize our findings below:
\vspace{-1ex}
\begin{itemize}[leftmargin=*]
\setlength\itemsep{0.1mm}
\item \textbf{Single-task vs. Multi-task:} Without video-language (VidL) pre-training (L1-4), we find that multi-task training greatly improves model performance in both cases, as it can take advantages of additional and diverse supervision. Combining multi-task finetuning with VidL pre-training renders slight performance drop (L5-8).
\item \textbf{Single shared head vs. Task-specific heads:} Results show that \modelname (single shared head) outperforms \modelbaseline (task-specific heads), with (L8 vs. L6, +5.9) or without (L2 vs. L4, +2.9) VidL pre-training. \modelname also saves more parameters (approximately 3H) from the additional 3 task-specific heads in \modelbaseline under multi-task finetuning. In addition, the performance drop of multi-task finetuning from single-task finetuning is less severe with the shared MLM head (-0.6 for \modelname vs. -1.6 for \modelbaseline) under VidL pre-training.
\end{itemize}

\Paragraph{Multi-task Variants with \modelname.} We also explore different multi-task variants with our unified framework \modelname in Table~\ref{table:task-specific}: ($i$) L8: the vanilla version without any task-specific design; ($ii$) L9: with human-readable task-specific prompts (\emph{e.g.}, ``\textit{is the video-text paired, true or false?}" for video text matching), which has shown promising results for language understanding~\cite{sanh2021multitask}; and ($iii$) L10: with learnable task-specific tokens (\emph{e.g.}, a special token \texttt{[VTM]} for video text matching), which is in analogy to the task prefixes in~\cite{VL-T5}. Different from observations in~\cite{sanh2021multitask,VL-T5}, both task-specific prompts and tokens do not show a clear advantage over the vanilla version. We conjecture the differences may be due to the weaker text encoder and the less diverse text prompts, which we leave as interesting directions for future study. 
Based on the above analyses, we simply extend the vanilla multi-task finetuning method from 4 to all 14 VidL benchmarks considered. 

\Paragraph{All-in-One.}  In Table~\ref{table:multi-task}, we finally attempt to answer the question with one model that can conquer all 14 downstream tasks simultaneously. We first establish the baseline performance by training single-task (ST) models with \modelname. We then report multi-task results with ($i$) a single set of parameters for all tasks (MT (all-in-one)); ($ii$) the best-performing checkpoint for each task (MT (best)); and ($iii$) with multi-task finetuning as 2nd stage pre-training and then finetune on each single task (MT$\rightarrow$ST), from the learned weights with MT (all-in-one). As the results show, MT$\rightarrow$ST achieves slightly better Meta-Ave across all settings. Surprisingly, the all-in-one model is very competitive, with only -0.5 performance drop on Meta-Ave, when compared with ST models.
\begin{table}[t]
\centering
    \tablestyle{2pt}{1.2} 
    \def \w{20pt} 
    \caption{\textbf{Multi-task Finetuning}. Accuracy, average (R1, R5, R10) and CIDEr score are used as evaluation metrics for video QA, retrieval and captioning tasks. Meta-Ave. is the average score across all evaluation datasets. P denotes the total parameter count in \modelname. All results are reported under VidL pre-training.
    }
    \vspace{1ex}
    \label{table:multi-task}
    \footnotesize
    \resizebox{\linewidth}{!}{
    \begin{tabular}{lc|ccccccccccccccc}
        % \toprule
        \shline
        Finetune  & ~ &  Meta & \multicolumn{3}{c}{TGIF}  & \multicolumn{4}{c}{MSRVTT}  & \multicolumn{3}{c}{LSMDC}  & \multicolumn{3}{c}{MSVD} & DiDeMo\\
        \cmidrule(lr){4-6} \cmidrule(lr){7-10} \cmidrule(lr){11-13} \cmidrule(lr){14-16} \cmidrule(lr){17-17} 
        Method & \# Params & Ave. & Act. & Trans. & Frame  & MC & QA & Ret & Cap & MC & FiB  & Ret & QA & Ret & Cap & Ret \\
        \shline
        ST & 14P & 73.9 & 95.8 & \textbf{99.1} & \textbf{72.2}  & \textbf{96.6} & \textbf{44.2} & \textbf{58.9} & 57.3 & 84.5 & \textbf{56.9}  & \textbf{39.8} & 54.4 &  67.6 & 139.4  &  \textbf{68.2}\\
        MT \scriptsize{(all-in-one)} & P & 73.4 & 95.8 & 98.0 & 70.7 & 93.9 & 44.1 & 56.3 & 57.1 & 85.3 & 56.5 & 39.4 & 53.4 & 69.2 & 141.1 &  66.1\\
        MT \scriptsize{(best)} & 14P & 73.8 & 95.8 & 98.3 & 71.6  & 94.3 & \textbf{44.2}  & 56.4 & 57.2 & \textbf{86.0} & 56.7 & 39.4  & \textbf{55.4}  & 69.3 & 141.6 & 66.5\\
        MT $\rightarrow$ ST & 14P & \textbf{74.2} & \textbf{96.6} & 98.5 &  71.2 & 96.0 & 44.1  & 58.8 & \textbf{58.0} & 85.3 & \textbf{56.9} & \textbf{39.8} & 53.5  & \textbf{69.7}  & \textbf{142.9} & 67.7\\
        \shline
    \end{tabular}
    }
    \vspace{-2ex}
\end{table}
\begin{figure*}[t!]
    \centering
    \begin{subfigure}[t]{0.24\textwidth}
        \centering
        \includegraphics[width=\textwidth]{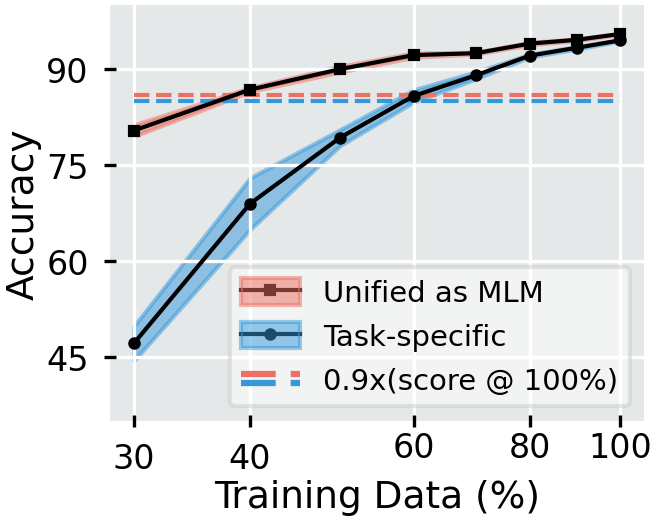}
        \caption{TGIF-Action}
    \end{subfigure}%
    ~ 
    \begin{subfigure}[t]{0.24\textwidth}
        \centering
        \includegraphics[width=\textwidth]{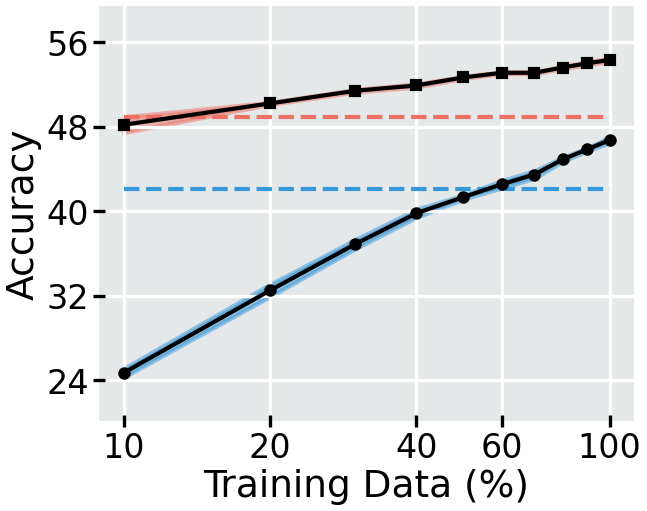}
        \caption{MSVD-QA}
    \end{subfigure}%
    ~ 
    \begin{subfigure}[t]{0.24\textwidth}
        \centering
        \includegraphics[width=\textwidth]{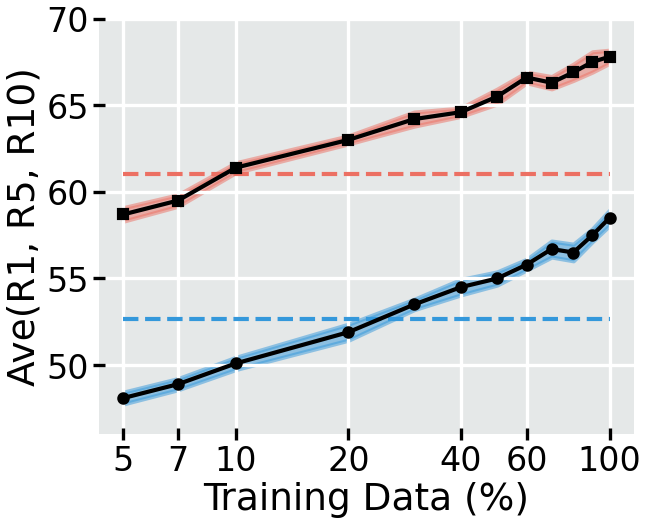}
        \caption{DiDeMo Retrieval}
    \end{subfigure}%
    ~ 
    \begin{subfigure}[t]{0.24\textwidth}
        \centering
        \includegraphics[width=\textwidth]{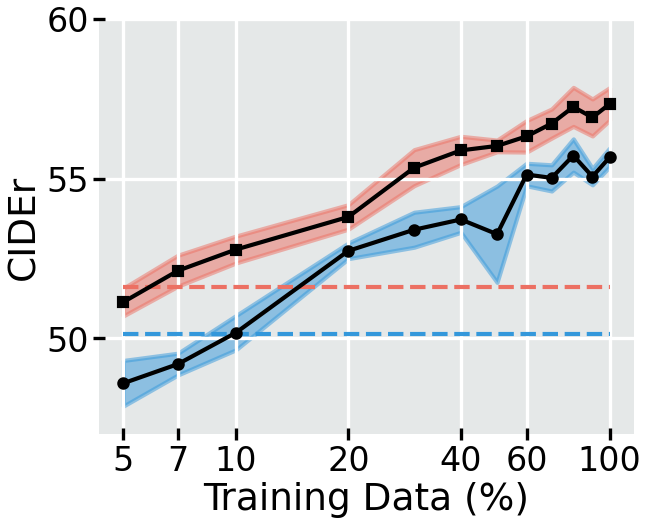}
        \caption{MSRVTT Captioning}
    \end{subfigure}
    \vspace{-1ex}
    \caption{\textbf{Few-shot Evaluation} under VidL Pre-training. Each experiment are repeated 5 times with different random seeds. The shaded areas highlight the standard error. Percentage of training data needed to achieve 90\% of the full model performance: (a) 40\%, (b) 10\%, (c) 10\%, (d) 6\% for \modelname (unified as MLM, red) and (a) 60\%, (b) 60\%, (c) 25\%, (d) 10\% for task-specific baseline \modelbaseline (blue).}
    \label{fig:few-shot}
    \vspace{-3ex}
\end{figure*}
\begin{table}[t]
\centering
    \tablestyle{5pt}{1.1} 
    \def \w{20pt} 
    
    \caption{\textbf{Zero-shot Evaluation on Video QA}. Top-1 accuracy is reported. We gray out BLIP~\cite{li2022blip}, which is additionally supervised with VQA v2~\cite{balanced_vqa_v2}, and \textsc{\footnotesize{Merlot Reserve}}~\cite{zellers2022merlot},  pre-trained with additional audio modality and uses GPT-3~\cite{gpt3} to reword questions into masked statements. $^\star$: unpublished preprints.}
    \label{table:zs}
    \vspace{1ex}
    \begin{tabular}{lc|cccccccc}
        \shline
        ~  & \# pre-train  & \multicolumn{3}{c}{TGIF}  & \multicolumn{2}{c}{MSRVTT}  & \multicolumn{2}{c}{LSMDC}  & MSVD \\
        \cmidrule(lr){3-5} \cmidrule(lr){6-7} \cmidrule(lr){8-9} \cmidrule(lr){10-10}
        Method & video/images  & Act. & Trans. & Frame  & MC & QA & MC & FiB   & QA \\
        \hline
        JustAsk~\cite{yang2021just-ask} & 69M / - & - &  - & - & - & 2.9 & - & - & 7.5\\
        \demph{\textsc{\footnotesize{Merlot Reserve}}~\cite{zellers2022merlot}} & \demph{1B/} - &  - & - & - & - & \demph{5.8}  & - & \demph{31.0} & - \\ 
        \demph{BLIP$^\star$}~\cite{li2022blip} & \demph{- / 129M} & - & - & - & - & \demph{19.2} & - & - & \demph{35.2}\\
        All-in-one$^\star$~\cite{all-in-one} & 283M / - & - & - & - & 80.3 & - & 56.3 & - & - \\
        \hline
        \modelbaseline & 2.5M / 3M & 48.5 & 47.9 & 0.0 & 84.6 & 0.0 & 66.9 & 0.0 & 0.0 \\
        \hline
        \multirow{2}{*}{\modelname} & 2.5M / 3M & 52.6 & \textbf{54.1} & 16.7 & 86.7 & \textbf{4.5} & 73.8 & 34.2 & \textbf{11.6}\\
        & 14M / 16M & \textbf{55.1} & 53.8 & \textbf{19.6} & \textbf{87.2} &  2.7 & \textbf{73.9} & \textbf{36.7} & 9.2 \\
        \shline
    \end{tabular}
    \vspace{-2ex}
\end{table}

\SubSection{Few-shot and Zero-shot Evaluation}\label{sec:zs-few-shot}
Next, we showcase two new capabilities enabled by \modelname over task-specific baseline. 

\Paragraph{Few-Shot Generalizability.} We first study how \modelname can generalize to testing data with limited training examples.  Figure~\ref{fig:few-shot} presents the results of the proposed \modelname (red line), which is unified as Masked Language Modeling (MLM) for all downstream tasks, in comparison to the task-specific baseline \modelbaseline with different heads and finetuning objectives (blue line) on 4 representative benchmarks. For easier comparison, we further plot two dotted lines, which denotes 90\% of model performance when trained with all the training data for each model. \modelname has shown a clear advantage, as it can easily achieve 90\% of the full model performance, with much less training data. Specifically, approximately 40\%, 10\%, 10\%, 6\% of training data is needed for \modelname on TGIF-Action, MSVD-QA, DiDeMo-Retrieval and MSRVTT-Captioning, while \modelbaseline requires 60\%, 60\%, 25\% and 10\%, respectively.

\Paragraph{Zero-Shot Evaluation on Video QA.} Even without a heavy text decoder as in ~\cite{VL-T5,unicorn,OFA}, a pre-trained \modelname can be directly evaluated on video question answering (QA) tasks without finetuning. Table~\ref{table:zs} compares zero-shot (ZS) performance of \modelname with task-specific baseline \modelbaseline on 8 video QA benchmarks.

Since the model has neither learned to perform the multiple-choice QA task nor seen similar data during pre-training, we transform multiple-choice QA as Video Text Matching (VTM) for better ZS performance. Specifically, we let \modelname to predict \texttt{true} or \texttt{false} via MLM head, given a video-question-answer input, and we rank the probability of model prediction as \texttt{true} across all answer choices. Similarly, for \modelbaseline with binary classification head for VTM, we simply rank the probability of model prediction as ``matched''. With the same pre-training data, \modelname evidently outperforms \modelbaseline on all multiple-choice QA benchmarks. On open-ended QA tasks, \modelname can be applied seamlessly, thanks to the shared MLM head. However, for \modelbaseline, the randomly initialized task-specific heads give meaningless ZS predictions. 

We also compare \modelname against previous methods. Without the help of additional audio modality in~\cite{zellers2022merlot} or supervision signals in~\cite{li2022blip}, \modelname achieves competitive ZS performance, even when pre-trained with much less data (5.5M vs. >69M). When scaling up the pre-training data by roughly 5 times, we observe notable performance improvements on most QA benchmarks. The performance drop on a few datasets may be due to the inclusion of more noisy data when scaling up. 
% These results suggest 
% that there is still room to boost model performance if we can further scale up the pre-training data, following the same recipe.

\SubSection{Comparison to Prior Arts}\label{sec:sota}
In this section, we compare \modelname with prior arts, which are mostly designed to tackle a single type of video-language (VidL) task. 

Table~\ref{table:qa} summarizes results of \modelname on video question answering (QA) and video captioning. For \textbf{video QA}, \modelname achieves significant gains over existing VidL pre-trained models on 7 out 8 video QA benchmarks considered. On MSRVTT-QA, \modelname is only 1.8 points behind All-in-one~\cite{all-in-one} pre-trained with 283M videos, which is 9 times more than ours (30M). It is worth mentioning that VIOLET~\cite{fu2021violet} adopts the same model architecture and finetuning objectives as our task-specific baseline \modelbaseline. Even when scaling up the VidL pre-training to 186M videos+images, the task-specific model VIOLET still underperforms \modelname, which further demonstrates the advantages of our unified framework. For \textbf{video captioning}, \modelname achieves the new state-of-the-arts on both datasets. Note that MV-GPT~\cite{MV-GPT} is pre-trained for multi-modal video captioning, where the auto-transcribed text from audio is used as additional input.  With video-only inputs, \modelname is able to achieve comparable performance. 

Table~\ref{table:retrieval-all} presents the comparison on \textbf{text-to-video retrieval}. The most competitive methods~\cite{camoe,qb-norm,bridgeformer} on text-to-video retrieval are based on CLIP~\cite{radford2021clip} pre-trained on 400M images. However, with much fewer pre-training data, \modelname can still perform competitively on all 4 benchmarks, especially when compared to non-CLIP pre-trained methods~\cite{all-in-one,fu2021violet,bain2021frozen}. Notably, on DiDeMo and LSMDC, \modelname surpasses all baseline methods in Table~\ref{table:retrieval-all}. We hypothesize that the fusion encoder in \modelname is more effective in modeling interactions between video and long paragraph query (\emph{i.e.}, DiDeMo) or the contextualized queries collected from movie scripts (\emph{i.e.}, LSMDC), than the late dot-product fusion in~\cite{camoe,qb-norm}.

\begin{table}[t]
\centering
    \tablestyle{3pt}{1.1} 
    \def \w{20pt} 
    \caption{Comparison with SOTA on \textbf{video question answering} and \textbf{video captioning}. Accuracy and CIDEr scores are reported for QA and captioning evaluation. 
    $^\star$: unpublished preprints.
    }
    \vspace{1ex}
    \label{table:qa}
    \begin{tabular}{ll|cccccccc|cc}
        \shline
        ~ & \# Pretrain  & \multicolumn{3}{c}{TGIF}  & \multicolumn{2}{c}{MSRVTT}  & \multicolumn{2}{c}{LSMDC}  & MSVD  & \multicolumn{2}{c}{Captioning}\\
        \cmidrule(lr){3-5} \cmidrule(lr){6-7} \cmidrule(lr){8-9} \cmidrule(lr){10-10} \cmidrule(lr){11-12}
        Method & videos/images & Act. & Trans. & Frame  & MC & QA  & MC & FiB  & QA & MSRVTT  & MSVD \\
        \shline
        ClipBERT~\cite{lei2021clip-bert} & - / 200K & 82.8 & 87.8 & 60.3  & 88.2 & 37.4  & - & -  & -  & -  & -\\
        JustAsk~\cite{yang2021just-ask}& 69M / -   & - & - & -  & - & 41.5  & - & -  & 46.3  & -  & -\\
        MERLOT~\cite{zellers2021merlot} & 180M / - &   94.0 & 96.2 & 69.5  & 90.9 & 43.1  & 81.7 & 52.9  & -  & -  & -\\
        VIOLET$^\star$~\cite{fu2021violet} & 183M / 3M & 92.5 & 95.7 & 68.9  & 91.9 & 43.9  & 82.8 & 53.7  & 47.9  & -  & -\\
        All-in-one$^\star$~\cite{all-in-one} & 283M / - & 95.5 & 94.7 & 66.3  & 92.3 &  \textbf{46.8} & 84.4 & -  & 48.3  & -  & -\\
        SwinBERT~\cite{lin2021swinbert} & - / - & - & - & -  & - & -  & - & -  & -  & 53.8  & 120.6\\
        MV-GPT~\cite{MV-GPT} & 53M / - & - & - & -  & - & 41.7 & - & -  & -  & 60.0  & -\\
         \hline
          \multirow{2}{*}{\modelname} & 2.5M / 3M & \textbf{96.6} & \textbf{99.1} & 72.2  & 96.6 & 44.2  & 86.0 & 56.9  & 55.4  & 58.0  & 142.9 \\
         & 14M / 16M & 96.3 & 98.7 & \textbf{73.5} & \textbf{97.4} & 45.0 & \textbf{87.0} & \textbf{57.1} & \textbf{56.6} & \textbf{60.1} & \textbf{150.7}\\
         \shline
    \end{tabular}
    \vspace{-3ex}
\end{table}

\begin{table}[t!]
\centering
    \tablestyle{3pt}{1.1} 
    \def \w{15pt}
    \caption{Comparison with SOTA on 
    \textbf{text-to-video-retrieval} tasks. All results are reported on R1/5/10. The row highlighted in blue assumes the model can see all textual queries during testing. $^\star$: unpublished preprints.
    }
    \vspace{1ex}
    \label{table:retrieval-all}
    \resizebox{\linewidth}{!}{
    \begin{tabular}{ll|cccc}
        \shline
        ~ & \# Pretrain & \multicolumn{4}{c}{Text-to-Video Retrieval} \\
        \cmidrule(lr){3-6}
        Method & videos/images & MSRVTT & DiDeMo & MSVD & LSMDC \\
        \shline
        ClipBERT~\cite{lei2021clip-bert} & - / 200K & 22.0 / 46.8 / 59.9 & 20.4 / 48.0 / 60.8 & - & - \\
        Frozen~\cite{bain2021frozen} & 2.5M / 3.2M  & 32.5 / 61.5 / 71.2 & 31.0 / 59.8 / 72.4 & 45.6 / 79.8 / 88.2 & 15.0 / 30.8 / 39.8 \\
        VIOLET$^\star$~\cite{fu2021violet}&  183M / 3M & 34.5 / 63.0 / 73.4 & 32.6 / 62.8 / 74.7 & - & 16.1 / 36.6 / 41.2 \\
        All-in-one$^\star$~\cite{all-in-one} & 103M / - & 37.9 / 68.1 / 77.1 & 32.7 / 61.4 / 73.5 & - & - \\
        BridgeFormer~\cite{bridgeformer} & - / 400M & 44.9 / 71.9 / 80.3 & - & \textbf{54.4 }/ \textbf{82.8 }/ 89.4 & 21.8 / 41.1 / 50.6 \\
        QB-Norm~\cite{qb-norm}& - / 400M & 47.2 / 73.0 / 83.0 & 43.3 / 71.4 / 80.8 &  47.6 / 77.6 / 86.1 & 22.4 / 40.1 / 49.5\\
        \rowcolor{LightCyan}
        CAMoE~\cite{camoe}& - / 400M & \textbf{47.3} /\textbf{ 74.2 }/ \textbf{84.5} & 43.8 / 71.4 / 79.9 & 49.8 / 79.2 / 87.0 & 25.9 / 46.1 / 53.7\\
         \hline
        \multirow{2}{*}{\modelname} & 2.5M / 3M & 37.8 / 63.8 / 75.0 & 47.4 / 74.7 / 82.4 & 46.3 / 76.9 / 86.0 & 22.2 / 43.8 / 53.5 \\
        & 14M / 16M & 40.7 / 66.9 / 77.6 & \textbf{53.4} / \textbf{78.6} / \textbf{85.3} & 50.1 / 79.6 / 87.2 & \textbf{26.1} / \textbf{46.4} / \textbf{57.3}\\
        \shline
    \end{tabular}
    }
    \vspace{-3ex}
\end{table}

\Section{Conclusion and Discussion of Broader Impact}\label{sec:conclusion}
We introduce \modelname, the first unified video-language (VidL) framework, that can tackle various VidL tasks with a unified Masked Language Modeling objective. Without any task-specific architectures, \modelname outperforms the prior state-of-the-art on 12 out of 14 benchmarks considered. Experiments show that \modelname is better suited for multi-task learning, few-shot generalization and zero-shot evaluation on video question answering tasks.
There are several potential limitations of \modelname that would make for promising directions for future work, including: ($i$) extension to fine-grained VidL tasks (\textit{e.g.}, video corpus moment retrieval~\cite{lei2020tvr}); and ($ii$) more effective in-context few-shot learning or prompt tuning. 
Like other data-driven systems, \modelname shares similar risks that may have negative societal impact, such as biases in training data and energy consumption with large-scale training. However, we believe that our unified framework combined with multi-task learning can most likely reduce both memory and energy costs, and potentially lead to more economical deployment in real-world applications.

\small{
\bibliographystyle{plain}
\bibliography{references} 
}

\clearpage

\appendix

\Section{Additional Results}\label{app:res}
\Paragraph{Multi-task Finetuning.} For completeness, we include additional results under multi-task finetuning. Table~\ref{table:multi-task_supp} presents the results of single-task finetuning and multi-task variants from the scale-up pre-training on 14M videos + 16M images. For easier reference in future work, we report detailed retrieval results on R1/5/10 in Table~\ref{table:supp_retrieval}.  

\begin{table}[h]
\centering
    \tablestyle{2pt}{1.2} 
    \def \w{20pt} 
    \caption{\textbf{Multi-task Finetuning of \modelname with scale-up VidL Pre-training on 14M videos + 16M images}. Accuracy, average(R1, R5, R10) and CIDEr score are used as evaluation metrics for video QA, retrieval and captioning tasks. Meta-Ave. is the average score across all evaluation datasets. P denotes the total parameter count in \modelname (backbone + MLM head).
    % $\dagger$ indicates the task-specific version of \modelname.
    }
    \vspace{1ex}
    \label{table:multi-task_supp}
    \footnotesize
    \resizebox{\linewidth}{!}{
    \begin{tabular}{lc|ccccccccccccccc}
        % \toprule
        \shline
        Finetune  & ~ &  Meta & \multicolumn{3}{c}{TGIF}  & \multicolumn{4}{c}{MSRVTT}  & \multicolumn{3}{c}{LSMDC}  & \multicolumn{3}{c}{MSVD} & DiDeMo\\
        \cmidrule(lr){4-6} \cmidrule(lr){7-10} \cmidrule(lr){11-13} \cmidrule(lr){14-16} \cmidrule(lr){17-17} 
        Method & \# Params & Ave. & Act. & Trans. & Frame  & MC & QA & Ret & Cap & MC & FiB  & Ret & QA & Ret & Cap & Ret \\
        \shline
        ST & 14P & 76.0 & 94.8 & 98.7 & 73.5 & 97.2 & \textbf{45.0} & 61.4 & 59.4 & 85.9 & \textbf{57.1} & 41.9 & 55.6 & \textbf{72.3} & 150.3 & \textbf{72.4}  \\
        MT \scriptsize{(all-in-one)} & P & 75.2 & 95.4 & 98.4 & 71.0 & 94.5 & 44.7 & 58.9 & 57.9 & \textbf{87.0} & 56.6 & 41.7  & 54.0 & 71.7  & 150.0 &  71.1\\
        MT \scriptsize{(best)} & 14P & 75.4 & 95.6 & 98.6 & 72.5 & 94.5 & 44.8 & 59.1 & 58.9 & \textbf{87.0} & 56.9 & 41.9 & \textbf{56.6} & 71.9 & 150.1 & 71.3\\
        MT $\rightarrow$ ST & 14P & 76.2 & \textbf{96.3} & 98.6 & 71.8 & \textbf{97.4}& \textbf{45.0} & \textbf{61.7} & \textbf{60.1} & \textbf{87.0} & \textbf{57.1} & \textbf{43.3} & 54.3 & 71.8 & \textbf{150.7} & 71.5\\
        \shline
        % \bottomrule
    \end{tabular}
    }
    \vspace{-2ex}
\end{table}

\begin{table}[h]
\centering
    \tablestyle{3pt}{1.2} 
    \def \w{15pt}
    \caption{ 
    \textbf{Detailed results of \modelname on text-to-video retrieval tasks} under single-task (ST) finetuning, and different multi-task (MT) finetuning settings. All results are reported on R1/5/10.
    }
    \vspace{1ex}
    \label{table:supp_retrieval}
    \begin{tabular}{ll|cccc}
        \shline
        \# Pretrain & Finetune & \multicolumn{4}{c}{Text-to-Video Retrieval} \\
        \cmidrule(lr){3-6}
       videos/images & Method &  MSRVTT & DiDeMo & MSVD & LSMDC \\
        \shline
       \multirow{4}{*}{2.5M / 3M} & ST &  \textbf{37.8} / \textbf{63.8} / 75.0 & \textbf{47.4} / \textbf{74.7} / 82.4 & 45.8 / 75.7 / 85.0 & 22.2 / \textbf{43.8} / \textbf{53.5} \\
        & MT \scriptsize{(all-in-one)}& 33.7 / 61.3 / 73.9  & 44.1 / 72.4 / 81.8  & 45.4 / 76.5 / 85.8 & 22.4 / 42.8 / 53.1\\
        & MT \scriptsize{(best)}& 34.9 / 61.1 / 73.2 & 46.3 / 72.2 / 81.1 & 46.2 / 76.2 / 85.5 & 22.4 / 42.8 / 53.1 \\
        & MT $\rightarrow$ ST & 36.8 / 63.4 / \textbf{75.2} & 45.9 / 73.0 / \textbf{84.1} &  \textbf{46.3} / \textbf{76.9} / \textbf{86.0} & \textbf{22.8} / 43.5 / 53.2\\
        \hline
        14M / 16M & ST & 39.7 / 66.7 / \textbf{77.8} & \textbf{53.4} / \textbf{78.6} / \textbf{85.3} & \textbf{50.1} / \textbf{79.6} / \textbf{87.2} & 25.1 / 44.6 / 56.0\\
        & MT \scriptsize{(all-in-one)}& 37.6 / 63.4 / 75.6 & 50.3 / 77.8 / 85.2 & 49.4 / 78.7 / 86.8  & 23.4 / 77.8 / 85.2\\
        & MT \scriptsize{(best)}& 37.5 / 64.2 / 75.7 & 51.0 / 78.1 / 84.7 & 49.9 / 79.0 / 86.8  &  24.1 / 46.3 / 55.4\\
        & MT $\rightarrow$ ST & \textbf{40.7} / \textbf{66.9} / 77.6 & 51.6 / 77.7 / 85.1 &  49.8 / 78.8 / 86.8 & \textbf{26.1} / \textbf{46.4} / \textbf{57.3}\\
        \shline
    \end{tabular}
    \vspace{-2ex}
\end{table}

\Paragraph{Investigation on Other Pre-training Tasks.}
As mentioned in the main text, we only adopt Masked Language Modeling (MLM) and Video Text Matching (VTM) as pre-training tasks for both the proposed \modelname and the task-specific baseline \modelbaseline. Here we briefly discuss other popular pre-training objectives with \modelbaseline. The first is \textbf{Frame Order Modeling}~\cite{li2020hero,zellers2021merlot}, where the input video frames are randomly shuffled and the goal is to revert back its original order. Different from the video-ASR pairs utilized in these works, the paired text in our pre-training data is not temporally grounded.  In most cases, the shuffled frame sequence will probably still be globally aligned with the textual description. Hence, such fine-grained temporal reasoning objective is not applicable in our case. The second is \textbf{Masked Visual Modeling} (MVM), where the model learns to reconstruct high-level semantics or low-level details for a certain percentage of ``masked'' visual inputs (\textit{i.e.}, features or patches). Different variants have been proposed and shown little-to-none effect in vision-language pre-training, such as predicting the object category of masked image regions~\cite{chen2020uniter} and distilling region/frame features from well-supervised vision encoders~\cite{chen2020uniter,li2020hero}. More recently, by taking advantage of pre-trained DALL-E~\cite{ramesh2021dalle}, researchers~\cite{bao2021beit,fu2021violet,tan2021vimpac} have shown potentials in masked visual token modeling, which asks the model to recover the discrete latent codes of the masked image patches. \cite{wei2021masked-feat} explores image feature descriptors such as Histograms of Oriented Gradients (HOG) as the prediction target for self-supervised visual pre-training. In Table~\ref{table:pretraining_tasks}, we investigate three different MVM objectives on top of VTM + MLM pre-training for \modelbaseline: ($i$) VQ Token: to recover the discrete codes extracted from pre-trained DALL-E following~\cite{fu2021violet}; ($ii$) Pixel: to regress the RGB colors as in~\cite{wei2021masked-feat}; and ($iii$) HOG: to regress the HOG values, following~\cite{wei2021masked-feat}. Results show that only MVM with HOG achieves a marginal performance improvement of +0.3 on average. Therefore, we adopt a simple recipe for all other pre-training experiments in the paper, that is with only MLM and VTM. 

\begin{table}[ht]
\centering
    \tablestyle{3pt}{1.2} 
    \def \w{20pt} 
    \caption{We investigate different \textbf{Masked Visual Modeling} tasks for pre-training \modelbaseline. Accuracy, average (R1, R5, R10) and CIDEr score are used as evaluation metrics for video QA, retrieval and captioning tasks. Meta-Ave. is the average score across all evaluation datasets. VidL pre-training is conducted on WebVid2.5M~\cite{bain2021frozen}.
    }
    \vspace{1ex}
    \label{table:pretraining_tasks}
    \begin{tabular}{l|cccccccccc}
        \shline
        Pre-raining   &  Meta & \multicolumn{3}{c}{TGIF}  & \multicolumn{3}{c}{MSRVTT}  & LSMDC  & MSVD & DiDeMo\\
        \cmidrule(lr){3-5} \cmidrule(lr){6-8}  \cmidrule(lr){9-9} \cmidrule(lr){10-10} \cmidrule(lr){11-11}
        Tasks  & Ave. & Act. & Trans. & Frame  & QA & Ret & Cap & FiB  & QA & Ret \\
        \shline
        VTM+MLM & 60.9 & 91.5 & \textbf{98.6} & \textbf{64.6} & \textbf{40.7} & 50.6 & 53.0 & \textbf{51.9} & 45.3 & 52.2 \\
        + VQ Token~\cite{fu2021violet} & 60.8 & \textbf{92.4} & \textbf{98.6} & 63.9 & 40.3 & \textbf{52.1} & 52.5 & 51.1 & 44.5 & 51.6 \\
        + Pixel~\cite{wei2021masked-feat} & 60.3 & 91.0 & 98.4 & 63.4 & 40.5 & 52.3 & 50.7 & 51.6 & 42.5 & 52.2 \\
        + HOG~\cite{wei2021masked-feat} & \textbf{61.2} & 91.7 & \textbf{98.6} & 64.4 & 40.4 & 51.8 & \textbf{53.4} & 50.5 &  \textbf{45.8} & \textbf{53.9}\\
        \shline
    \end{tabular}
    \vspace{-2ex}
\end{table}

\begin{figure*}[t!]
    \centering
    \begin{subfigure}[t]{\textwidth}
        \centering
        \includegraphics[width=\textwidth]{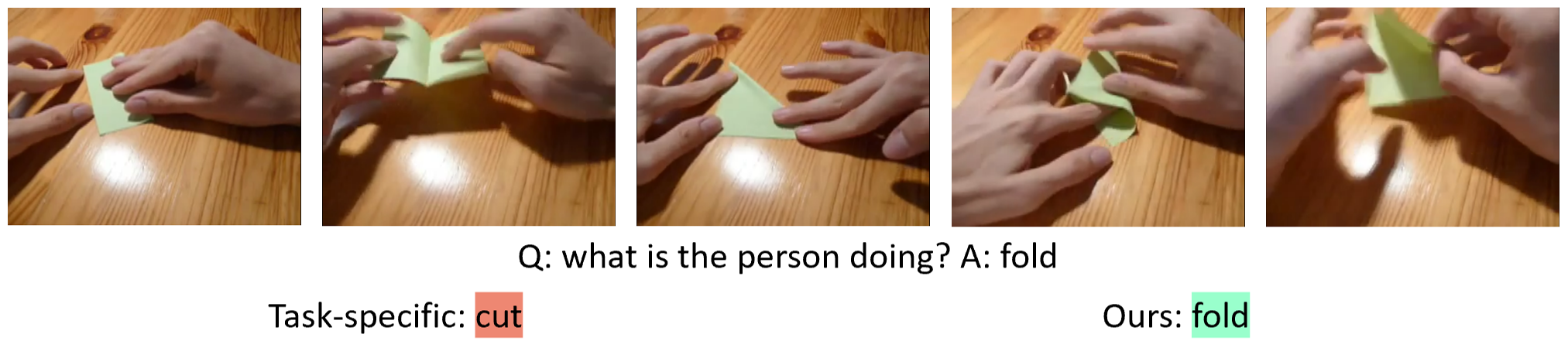}
        \caption{}
        \label{subfig:qa_1}
    \end{subfigure}
    \begin{subfigure}[t]{\textwidth}
        \centering
        \includegraphics[width=\textwidth]{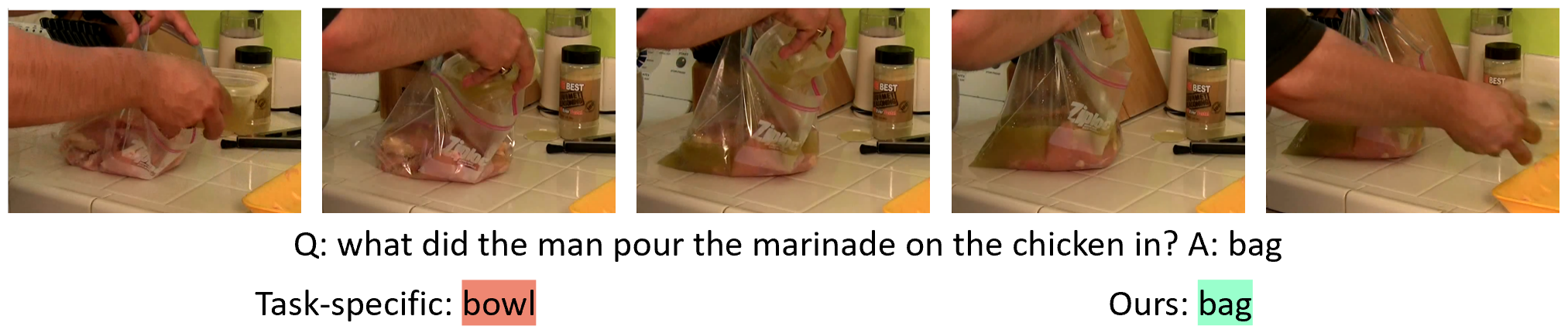}
        \caption{}
        \label{subfig:qa_2}
    \end{subfigure}
    \begin{subfigure}[t]{\textwidth}
        \centering
        \includegraphics[width=\textwidth]{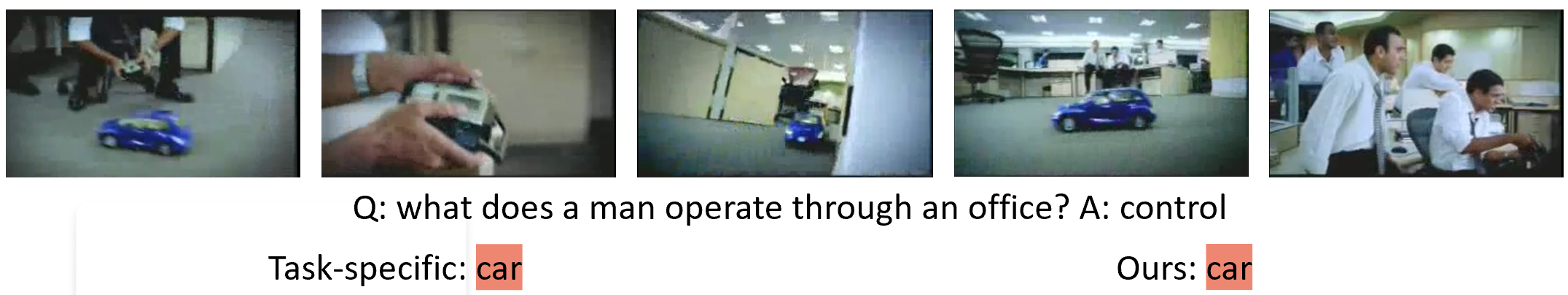}
        \caption{}
        \label{subfig:qa_3}
    \end{subfigure}
    \vspace{-1ex}
    \caption{\textbf{Visualization of model predictions from  \modelname (ours) and \modelbaseline (task-specific)} on video question answering. Green (Red) highlights the correct (wrong) predictions. 
    }
    \label{fig:supp_vis}
    \vspace{-2ex}
\end{figure*}

\Paragraph{Qualitative Comparisons to Task-specific Baseline.} Figure~\ref{fig:supp_vis} provides qualitative comparisons between \modelname and task-specific baseline \modelbaseline on video question answering (QA). The model predictions are sampled from MSVD-QA. 

In Figure~\ref{subfig:qa_1}, the ground-truth answer ``fold'' is not in the top-$k$ ($k=1000$ for MSVD-QA, following~\cite{fu2021violet}) most common answers in training split, hence excluded from the pre-defined answer vocabulary for training \modelbaseline. In Figure~\ref{subfig:qa_2}, the ground-truth answer ``bowl'' appears roughly 9 times more than ``bag" in the training split. These visualization results on video QA suggest that ($i$) our \modelname can better fit the open-ended setting for QA tasks, as it does not restrict the predictions to be from a pre-defined answer vocabulary as in \modelbaseline  (Figure~\ref{subfig:qa_1});  and ($ii$) the task-specific baseline is easier to fail on questions with out-of-distribution answers than \modelname (Figure~\ref{subfig:qa_2}). Additionally, we show in Figure~\ref{subfig:qa_3} when both models can provide reasonable answers to the question, which do not exactly match the ground-truth answer. This result reveals potential problems with the current evaluation metrics or existing datasets on video QA. Future work may consider collecting additional annotations to enrich the dataset and improve the evaluation metric to handle multiple ground-truth answers (\textit{e.g.}, similar to VQA scores~\cite{balanced_vqa_v2}).

\Section{Additional Comparison with Existing Work}

\begin{figure*}[t!]
    \centering
    \begin{subfigure}[t]{0.28\textwidth}
        \centering
        \includegraphics[width=\textwidth]{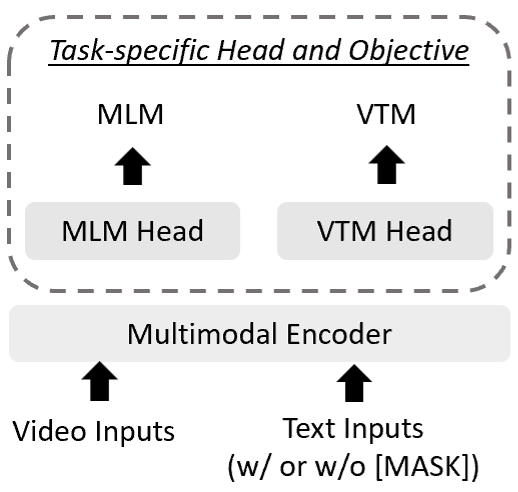}
        \caption{Existing VidL Methods}
        \label{subfig:existing_pretrain}
    \end{subfigure}%
    \hfill 
    \begin{subfigure}[t]{0.38\textwidth}
        \centering
        \includegraphics[width=0.73\textwidth]{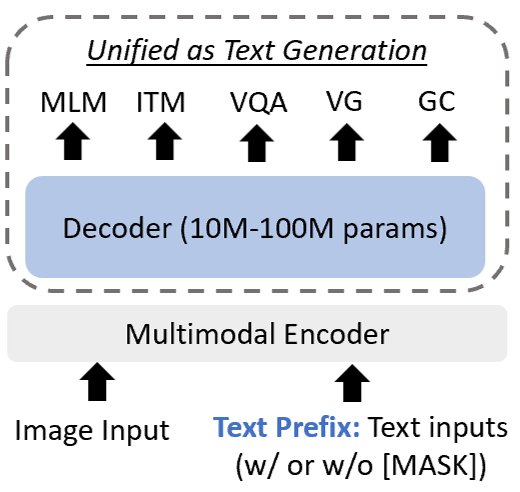}
        \caption{\footnotesize Existing Unified Image-text Models}
        \label{subfig:vl-t5_pretrain}
    \end{subfigure}%
    \hfill 
    \begin{subfigure}[t]{0.28\textwidth}
        \centering
        \includegraphics[width=\textwidth]{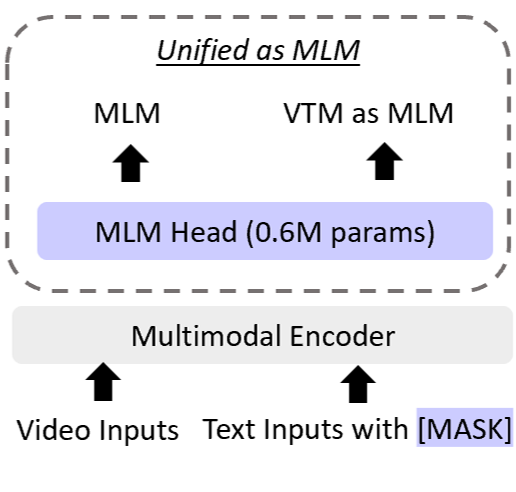}
        \caption{\modelname}
        \label{subfig:ours_pretrain}
    \end{subfigure}%
    \vspace{-1ex}
    \caption{Illustration of the \textbf{differences between \modelname and existing methods} during pre-training. \modelname unifies both masked language modeling (MLM) and video text matching (VTM) as MLM, without task-specific heads in existing video-language (VidL) models. Take VL-T5~\cite{VL-T5} as an example; most unified image-text models are pre-trained with a combination of complex pre-training tasks (\textit{e.g.}, visual question answering (VQA), visual grounding (VG), grounded captioning (GC)). 
    }
    \label{fig:comparison_pretrain}
    \vspace{-2ex}
\end{figure*}
Figure 2 in the main text shows the detailed comparison between \modelname and existing methods with the image/video question answering task as an example. In Figure~\ref{fig:comparison_pretrain}, we illustrate the differences among these methods in pre-training. Unlike existing video-language models, which design task-specific heads and objectives for different pre-training tasks.  \modelname unifies masked language modeling (MLM) and video text matching (VTM) as MLM. Compared with unified image-text models (\textit{e.g.}, VL-T5~\cite{VL-T5}), which are typically pre-trained with a combination of complex pre-training tasks, such as visual question answering and grounded captioning. Although these pre-training tasks may enable the model with new abilities (\textit{e.g.}, generating region proposals as in \cite{unicorn,OFA}), the supervision often comes from human-annotated data. It remains unclear how to design and effectively pre-train a unified model with such capability but without dependency on human-labeling.

\Section{Implementation Details}\label{app:implementation}
\Paragraph{Task-specific Prompts and Tokens.} As mentioned in Section 4.3 of the main text, we explore the vanilla multi-task finetuning without any task-specific designs and two additional variants with task-specific prompts and tokens for \modelname. Here, we describe the prompts and tokens used in these baselines.

For \textbf{task-specific prompts}, we insert a human-readable text prompt at the beginning of the text input. 
\begin{itemize}[leftmargin=*]
    \item For text-to-video retrieval and video-text matching during pre-training, the text prompt is \textit{``is the video-text paired, true or false''};
    \item For multiple-choice video question answering (QA), the text prompt is \textit{``which answer choice is correct, choose from 0, 1, 2, 3, 4.''};
    \item For open-ended video QA, the text prompt is \textit{``answer the question about the video.''};
    \item For video captioning, the text prompt is \textit{``write a description about the video.''}.
\end{itemize}
As discussed in the Experiments section of the main text, we only briefly investigate prompt tuning with \modelname. How to design more diverse prompts for more effective prompt tuning is an interesting direction for future work. 

For \textbf{task-specific tokens}, we add several new tokens to the whole vocabulary, and learn these token embeddings from scratch during multi-task finetuning. For both training and inference, the task-specific token is inserted right after \texttt{[CLS]} token in the text input. Specifically, the new task-specific tokens are \texttt{[VTM]} for text-to-video retrieval and video-text matching, \texttt{[MC]} for multi-choice video QA, \texttt{[OE]} for open-ended video QA, \texttt{[CAP]} for video captioning.

\Paragraph{Additional Training Details.} We summarize the training configurations for downstream finetuning in Table~\ref{table:supp_finetuning}. Due to various data scales and domains, we use task-specific batch size and training epochs based on the performance of the validation set for each downstream task (Table~\ref{subtab:task-specific}). All other settings are shared across all datasets (Table~\ref{subtab:common}). All experiments are conducted on Microsoft Azure~\cite{msft-azure}, adopting mixed-precision training with DeepSpeed~\cite{rasley2020deepspeed}. All video data are pre-processed by evenly extracting 32 frames to avoid expensive decoding on-the-fly. During training, we randomly sample $T$ frames from 32 frames, resize the shorter side of all frames to 224 and random crop (224x224) at the same location for all the frames in a given video. During inference, we evenly sample $T$ frames from 32 frames and center crop (224x224) for all the frames. 

For multi-task finetuning, since the same set of videos are shared among several downstream tasks, there might be overlaps between one's training split and others' validation or testing split (\textit{e.g.}, some video-text pairs in MSRVTT-Retrieval 9K-train is in the testing split of MSRVTT-Captioning). To avoid data contamination, we filter out validation and testing videos in all downstream datasets from the training splits, and use this cleaned version for multi-task finetuning. At each training step, we randomly sample one dataset from all 14 of them, and construct a batch of examples from that dataset. The training is conducted on 16$\times$80GB A100 for 20 epochs, and we adopt the same batch size for retrieval tasks as shown in Table~\ref{subtab:task-specific}, and batch size 60 for all other tasks.

\begin{table}[ht]
\centering
\caption{Training Configurations For Downstream Finetuning.}
\label{table:supp_finetuning}
\subfloat[Common Configurations. 
\label{subtab:common}]{
    \tablestyle{2pt}{1.2} 
    \begin{tabular}{lc}
        \shline
        Learning Rate & 2e-5 \\
        Weight Decay & 1e-3 \\
        Optimizer & AdamW~\cite{loshchilov2019adamw} \\
        $\beta$s & (0.9, 0.98) \\
        Warmup Ratio & 10\% \\
        \# Frames ($T$) & 5 \\
        Frame Size ($H$, $W$) & (224, 224) \\
        Patch Size ($h$, $w$) & (32, 32) \\
        \shline
\end{tabular}
}
\hfill
\subfloat[Task-specific Configurations. 
\label{subtab:task-specific}]{
    \tablestyle{2pt}{1.2} 
    \label{table:task_config}
    \begin{tabular}{l|ccc}
        % \toprule
        \shline
        Dataset & \# GPUs  & Batch Size / GPU & \# Epochs \\
        \shline
        \multicolumn{3}{l}{\textit{Video Question Answering}}\\
        \hline
        TGIF-Action   & \multirow{7}{*}{8$\times$32GB V100} & \multirow{7}{*}{24} & 56\\
        TGIF-Transition &  &  & 15 \\
        TGIF-Frame &  & & 10\\
        MSRVTT-QA & & & 8 \\
        LSMDC-MC &  & & 10 \\
        LSMDC-FiB & & & 5\\
        MSVD-QA & & & 8\\
        \hline
        \multicolumn{3}{l}{\textit{Text-to-Video Retrieval}}\\
        \hline
        MSRVTT & 16$\times$80GB A100 & \multirow{3}{*}{20} & 10\\
        \cmidrule(lr){2-2}
        LSMDC & \multirow{3}{*}{8$\times$80GB A100} &  & 5\\
        MSVD & & & 5\\
        \cmidrule(lr){3-3}
        DiDeMo & & 16 & 10 \\
        \hline
        \multicolumn{3}{l}{\textit{Video Captioning}}\\
        \hline
        MSRVTT & \multirow{2}{*}{8$\times$32GB V100} & \multirow{2}{*}{24} & \multirow{2}{*}{20} \\
        MSVD & & & \\
        \shline
    \end{tabular}
}
\end{table}
\Section{Pre-training Data}

\Paragraph{Public Datasets} We use the following publically available datasets to pretrain \modelname:
\begin{itemize}[leftmargin=*]
    \item \textbf{WebVid2.5M}~\cite{bain2021frozen} scrapes 2.5M video-text pairs from the web. The texts in this data are alt-text descriptions, which generally describe the global video semantics. 
    \item \textbf{Conceptual Captions 3M} (CC3M) ~\cite{sharma2018cc} consists of 3.3M image-text pairs, which are also harvested from the web. \textbf{CC12M}~\cite{changpinyo2021cc12m} further enlarges CC3M by 4 times. Both have been used to pre-train large-scale image-text models.
    \item \textbf{SBU-Captions}~\cite{sbu} is another widely used dataset for image-text pre-training, web-crawled from Flickr. It contains 1M image-text pairs. 
    \item \textbf{COCO}~\cite{chen2015coco} and \textbf{Visual Genome} (VG) are two human-annotated image-text datasets. COCO contains 5 captions per image over 120K images. Unlike COCO captions that can describe the whole scene, VG collects 5M regional descriptions over more than 100K images.
\end{itemize}

\Paragraph{Video-Text Data Collection} For the scale-up pre-training, we additionally crawl 11.9M video-text pairs from the web, following the same procedure in~\cite{bain2021frozen}. Here, we briefly describe how we collected the data.

WebVid2.5M has led to promising results in text-to-video retrieval tasks as shown in~\cite{bain2021frozen}. This motivates us to further crawl more video-text pairs from the same source. We first use a search engine to identify the potential data sources based on sampled textual descriptions in WebVid2.5M, and then we scrape the video-text pairs from these data sources.
Similarly, we follow \cite{bain2021frozen,sharma2018cc} to filter out offensive content and hide person and location names.
In total, we have collected 11.9M videos, each accompanied with an alt-text description.  The collected dataset shares similar characteristics as WebVid2.5M, with the average video duration as $\sim$ 20 seconds, and the average number of words in the textual description as $\sim$20. Note that at the time when we started this project, WebVid10M in \cite{bain2021frozen} has not been released yet. We later found our 11.9M data largely overlaps with WebVid10M. Hence, we refer future work to WebVid10M for scale-up pre-training.

\Section{Downstream Datasets}\label{app:downstream_dataset}
In this section, we introduce all downstream datasets used for evaluating \modelname and discuss some dataset-specific training details below. Table~\ref{table:supp_dataset} summarizes the number of examples in training/validation/testing split for each dataset.
\begin{table}[h]
\centering
    \tablestyle{3pt}{1.2} 
    \def \w{20pt} 
    \caption{\textbf{Data Distribution of Downstream Datasets}.
    }
    \label{table:supp_dataset}
\begin{minipage}{\linewidth}
\centering
\subfloat[Video Question Answering Tasks (\# videos / \# video-question pairs). For open-ended QA, we do not restrict the answer vocabulary to contain only the most common answers in training split. Theoretically, the model predictions can be any word in the whole vocabulary of \texttt{vocab\textunderscore size} = 30,522.
\label{subtab:qa_dist}]{
    \tablestyle{3pt}{1.2} 
    \resizebox{\linewidth}{!}{
    \begin{tabular}{lcccccccc}
        \shline
        & \multicolumn{3}{c}{TGIF}  & \multicolumn{2}{c}{MSRVTT}  & \multicolumn{2}{c}{LSMDC}  & MSVD \\
        \cmidrule(lr){2-4} \cmidrule(lr){5-6} \cmidrule(lr){7-8} \cmidrule(lr){9-9}  
          & Action & Transition & Frame  & QA & MC  & FiB  & MC & QA  \\
        \shline
        Training & 18K / 18K & 26K / 47K & 30K / 35K & 6.5K / 149K & - / - & 95K / 297K &  101K / 101K & 1.2K / 30K\\
        Validation & 2K / 2K & 5K / 5K &  4K/ 4K&  0.5K / 123K & - / - & 7K / 22K & 7K / 7K & 0.2K / 6K\\
        Testing & 2K / 2K & 3K / 6K&  7K/ 14K & 3K / 73K & 3K / 3K & 9.5K / 30K & 10K/ 10K & 0.5K / 13K\\
        \# \scriptsize{answer choices}  & \multirow{2}{*}{5} & \multirow{2}{*}{5} & \multirow{2}{*}{-} & \multirow{2}{*}{-}  & \multirow{2}{*}{5} & \multirow{2}{*}{-}  & \multirow{2}{*}{5} & \multirow{2}{*}{-} \\ 
       (MC-QA) & & & & & & & \\
        \shline
\end{tabular}
}
}
\end{minipage}
\begin{minipage}{\linewidth}
\centering
\subfloat[Text-to-video Retrieval and Video Captioning Tasks (\# videos / \# video-text pairs).$^\dagger$: on MSRVTT-Retrieval, we use the same split of 9K-training in~\cite{luo2021clip4clip,bain2021frozen}. For the validation purpose, we use the original validation split in 7K-training version, whose examples are included in the training split of 9K-training.  \label{subtab:ret_dist}]{
    \tablestyle{8pt}{1.2} 
    \begin{tabular}{lcccccc}
        \shline
        & \multicolumn{2}{c}{MSRVTT}  & \multicolumn{2}{c}{MSVD}  & LSMDC  &  DiDeMo \\
        \cmidrule(lr){2-3} \cmidrule(lr){4-5} \cmidrule(lr){6-6} \cmidrule(lr){7-7}  
          & Ret. & Cap. & Ret.  & Cap. & Ret  & Ret \\
        \shline
        Training & 9K / 180K & 6.5K / 130K &  1.2K / 49K  &  1.2K / 49K & 101K / 101K & 8K / 8K\\
        Validation & 1K / 1K$^\dagger$  & 0.5K / 10K &  0.1K / 4K   &  0.1K / 4K  & 7K / 7K & 1K / 1K \\
        Testing & 1K / 1K & 3K / 60K &  0.7K / 28K  &  0.7K / 28K & 1K / 1K & 1K / 1K\\
        \shline
\end{tabular}
}
\end{minipage}
\end{table}

\Paragraph{Text-to-video Retrieval} We evaluate \modelname on 4 popular text-to-video retrieval datasets, namely MSRVTT~\cite{xu2016msrvtt}, DiDeMo~\cite{hendricks2017didemo}, MSVD~\cite{chen2011msvd} and LSMDC~\cite{rohrbach2015lsmdc}.
\textbf{MSRVTT} contains 10K YouTube videos with 200K descriptions. We follow~\cite{bain2021frozen} to train on 9K videos and evaluate on 1K-A testing split. \textbf{DiDeMo} consists of 10K Flickr videos, each annotated with 4 sentences. We concatenate all sentences from the same video into a paragraph and perform paragraph-to-video retrieval, following~\cite{lei2021clip-bert, bain2021frozen}. Although this dataset comes with localisation annotations (ground-truth temporal proposals) for each sentence, we perform all experiments without leveraging this fine-grained information for both training and evaluation. Instead, we use the same procedure as described in Appendix~\ref{app:implementation} to sample frames from videos.  \textbf{MSVD} is based on 2K YouTube videos and crowdsourced 40 textual descriptions per video. \textbf{LSMDC} is built upon 118K video clips from 202 movies. Each clip has a caption from movie scripts or descriptive video services. We use the standard splits for DiDeMo, MSVD and LSMDC, following~\cite{luo2021clip4clip}.  For paragraph-to-video retrieval on DiDeMo,
we adopt the text augmentation technique proposed in Frozen~\cite{bain2021frozen}, which is to randomly sample and concatenate a variable number of sentences as paragraph for each video.

\Paragraph{Multiple-choice Video QA} We evaluate \modelname on four multiple-choice QA datasets: TGIF-Action, TGIF-Transition~\cite{jang2017tgif-qa}, MSRVTT-MC~\cite{xu2017msrvtt-msvd-qa} and LSMDC-MC~\cite{lsmdc-fib}. Among them, \textbf{TGIF-Action} and \textbf{TGIF-Transition} aim to test the model's ability to recognize repeating actions and state transitions in short GIFs. Each video-question pair is accompanied with 5 answer choices. We concatenate the 5 answer choices sequentially with the question, and the model is asked to predict the ground-truth answer index.  \textbf{MSRVTT-MC} and \textbf{LSMDC-MC} are based on retrieval tasks, but reformulated as multiple-choice QA. A model needs to find the caption that describes the video out of 5 candidate captions. Due to its similarity to video-to-text retrieval, we formulate it as video-text matching, which is the same as zero-shot evaluation described in the Experiments section of the main text.  Specifically, we let \modelname to predict \texttt{true} or \texttt{false} via MLM head, given a video-question-answer input, and we rank the probability of model prediction as \texttt{true} across all answer choices. As there is no training and validation data constructed in the same way for MSRVTT-MC, we follow~\cite{lei2021clip-bert} to evaluate the retrieval model trained on MSRVTT to rank the 5 candidate answers.

\Paragraph{Open-ended Video QA} Four datasets are considered for open-ended video QA: TGIF-Frame~\cite{jang2017tgif-qa}, MSRVTT-QA, MSVD-QA~\cite{xu2017msrvtt-msvd-qa} and LSMDC-FiB~\cite{lsmdc-fib}. Among them, the question-answer pairs in all but TGIF-Frame are based on the linguistic transformation of captions for each video. Questions in \textbf{TGIF-Frame} is collected via crowd-sourcing, which are answerable with just a single frame in the video. \textbf{MSRVTT-QA} contains 243K open-ended questions over 10K videos and \textbf{MSVD-QA} consists of 47K questions over 2K videos. The Fil-in-the-blank (FiB) task of \textbf{LSMDC-FiB} is, given a video and a sentence with a blank in it, to predict a correct word for the blank. We replace the blank with a \texttt{[MASK]} token, and naturally it becomes a Masked Language Modeling (MLM) task.

As mentioned in the main text, \modelname answers the open-ended questions in these datasets with only one word, as there is only one \texttt{[MASK]} token appended to the text input. Table~\ref{table:supp_oe_ans} summarizes the max answer length and the percentage of examples with answers longer than one word in all four datasets. As the statistics show, more than 92\% of examples in these datasets are answerable with a single word.
\begin{table}[h]
\centering
    \tablestyle{2pt}{1.2} 
    \def \w{20pt} 
    \caption{\textbf{Answer Length Distribution for Open-ended Video Question Answering}. In summary, there are < 8\% of examples across training, validation and testing split of each dataset, with answer length > 1. 
    }
    \label{table:supp_oe_ans}
    \vspace{1ex}
    \begin{tabular}{l|cccccccccccc}
        \shline
       ~  & \multicolumn{3}{c}{TGIF-Frame}  & \multicolumn{3}{c}{MSRVTT-QA}  & \multicolumn{3}{c}{LSMDC-FiB}  & \multicolumn{3}{c}{MSVD-QA}\\
        \cmidrule(lr){2-4} \cmidrule(lr){5-7}  \cmidrule(lr){8-10} \cmidrule(lr){11-13}
        ~ & Train & Val & Test & Train & Val & Test  & Train & Val & Test & Train & Val & Test  \\
        \shline
       Max answer\textunderscore len & 4 & 5 & 5 & 4 & 7& 6 & 3 & 3 & 3 & 4 & 4 & 4\\
       \% of data w/ answer\textunderscore len > 1 & 2.8 & 2.8 & 2.5 & 2.9 & 4.9& 4.6 & 2.4 & 2.7 & 2.6 & 6.6 & 7.9 & 7.0 \\
        \shline
    \end{tabular}
    \vspace{-2ex}
\end{table}

\Paragraph{Video Captioning} MSRVTT~\cite{xu2016msrvtt} and MSVD~\cite{chen2011msvd} are used for captioning evaluation. As introduced before, \textbf{MSRVTT} consists of 10K videos with 20 captions per video, and \textbf{MSVD} contains 2K videos, with 40 captions per video. We follow the standard captioning splits in \cite{xu2016msrvtt} and \cite{venugopalan2014translating} for MSRVTT and MSVD, respectively. During training, we set the probability of random masking caption tokens to be 0.15, the same as what is used in MLM during pre-training. During inference, we perform generation until the model outputs a \texttt{[SEP]}, which is defined as the sentence ending token or when it reaches the maximum generation step 50. 

\section{Data Usage}\label{sec:content_license_usage}
We strictly comply with the data usage agreement for each public dataset used in this paper, which is to only use for non-commercial research purposes. To the best of our knowledge, there is no personal identification information or offensive content in any of the public datasets.

\end{document}